\documentclass[10pt,twocolumn,letterpaper]{article}

\usepackage[pagenumbers]{wacv} 

\usepackage{graphicx}
\usepackage{amsmath}
\usepackage{amssymb}
\usepackage{booktabs}

\usepackage[utf8]{inputenc}
\usepackage[table]{xcolor}  
\usepackage{hhline}
\usepackage{adjustbox}
\usepackage{multirow}
\usepackage{array}

\usepackage{type1cm}
\usepackage{indentfirst}
\usepackage{graphicx}
\usepackage{bm}
\usepackage{amsfonts}
\usepackage{enumitem}
\usepackage[pagebackref,breaklinks,colorlinks]{hyperref}
\usepackage[capitalize]{cleveref}
\crefname{section}{Sec.}{Secs.}
\Crefname{section}{Section}{Sections}
\Crefname{table}{Table}{Tables}
\crefname{table}{Tab.}{Tabs.}

\def\wacvPaperID{329} 
\def\confName{WACV}
\def\confYear{2025}

\usepackage{fancyhdr}
\fancypagestyle{firstpage}{
  \fancyhf{}                    
  
  \fancyhead[C]{%
    \small
    This paper has been accepted to WACV 2025\\
    February 28--March 4, 2025\\
    JW Marriott Starr Pass Resort, Tucson, Arizona, USA
  }
  \fancyfoot[C]{\thepage}     
}

\begin{document}

\title{Decoupled PROB: Decoupled Query Initialization Tasks and Objectness-Class Learning for Open World Object Detection}

\author{
    Riku Inoue \hspace{1em} Masamitsu Tsuchiya \hspace{1em} Yuji Yasui \\
    Honda R$\&$D Co., Ltd. \\
    {\tt\small riku\_inoue@jp.honda}
}
\maketitle
\thispagestyle{firstpage} 
\begin{abstract}
Open World Object Detection (OWOD) is a challenging computer vision task that extends standard object detection by 
(1) detecting and classifying unknown objects without supervision, and (2) incrementally learning new object classes without forgetting previously learned ones. 
The absence of ground truths for unknown objects makes OWOD tasks particularly challenging.
Many methods have addressed this by using pseudo-labels for unknown objects. 
The recently proposed \textbf{Pr}obabilistic \textbf{Ob}jectness transformer-based open-world detector (PROB) is a state-of-the-art model that does not require pseudo-labels for unknown objects, as it predicts probabilistic objectness. 
However, this method faces issues with learning conflicts between objectness and class predictions. 

To address this issue and further enhance performance, we propose a novel model, Decoupled PROB.
Decoupled PROB introduces Early Termination of Objectness Prediction (ETOP) to stop objectness predictions at appropriate layers in the decoder, resolving the learning conflicts between class and objectness predictions in PROB. 
Additionally, we introduce Task-Decoupled Query Initialization (TDQI), which efficiently extracts features of known and unknown objects, thereby improving performance.
TDQI is a query initialization method that combines query selection and learnable queries, and it is a module that can be easily integrated into existing DETR-based OWOD models. 
Extensive experiments on OWOD benchmarks demonstrate that Decoupled PROB surpasses all existing methods across several metrics, significantly improving performance.
\end{abstract}

\section{Introduction}

\begin{figure}[tp]
    \centering
    \begin{minipage}[t]{.32\hsize}
        \centering
        \includegraphics[width=\linewidth]{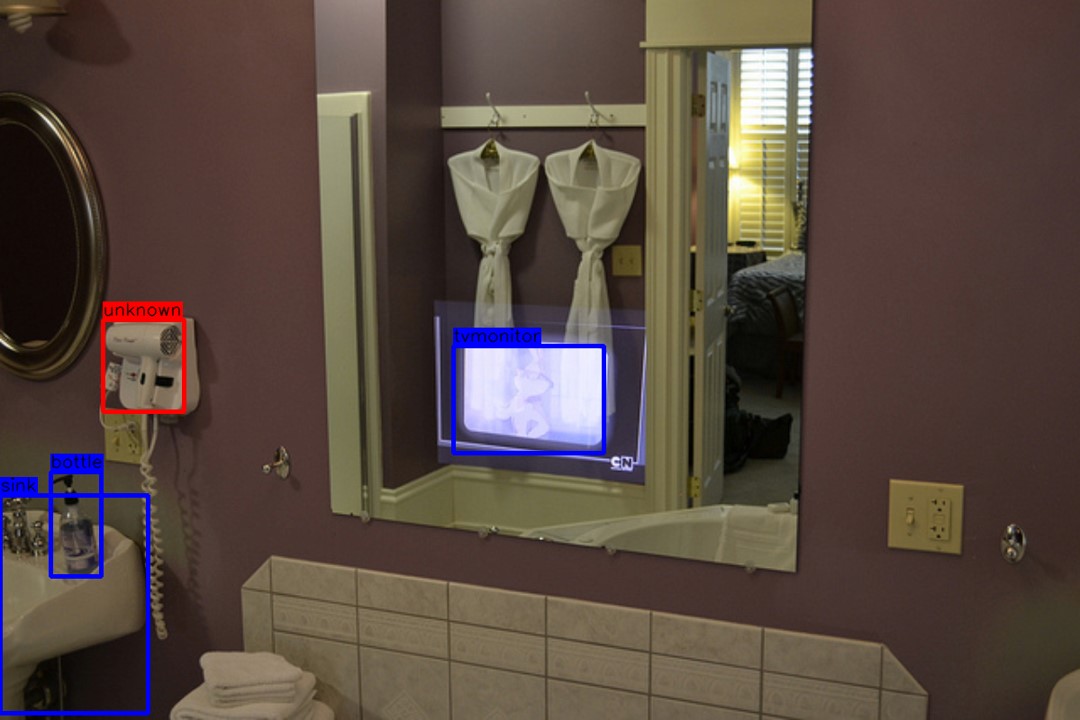}
        \label{fig:gt1}
    \end{minipage}
    \begin{minipage}[t]{.32\hsize}
        \centering
        \includegraphics[width=\linewidth]{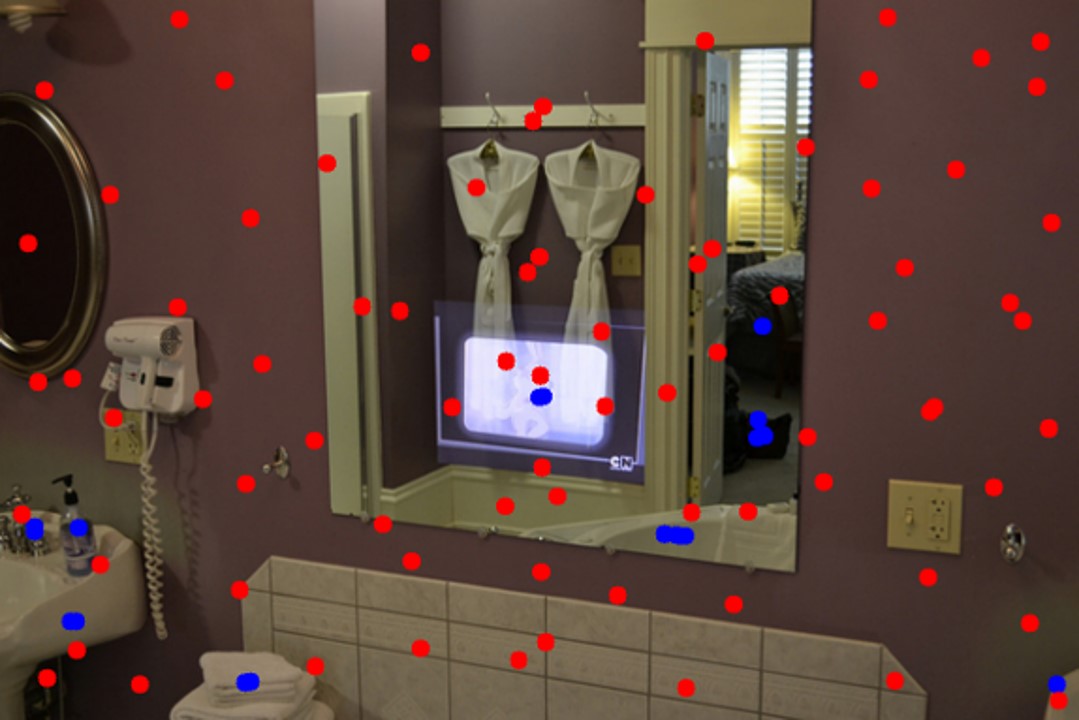}
        \label{fig:initref1}
    \end{minipage} 
    \begin{minipage}[t]{.32\hsize}
      \centering
      \includegraphics[width=\linewidth]{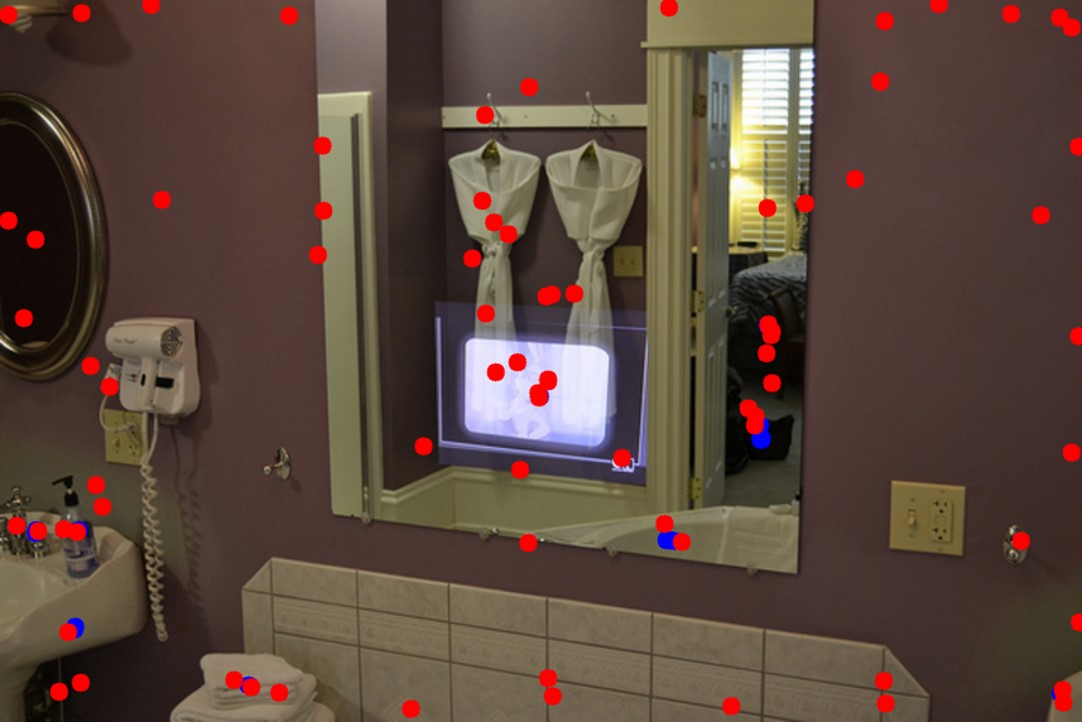}
      \label{fig:lastref1}
    \end{minipage} \\
  
    \centering
    \begin{minipage}[t]{.32\hsize}
        \centering
        \includegraphics[width=\linewidth]{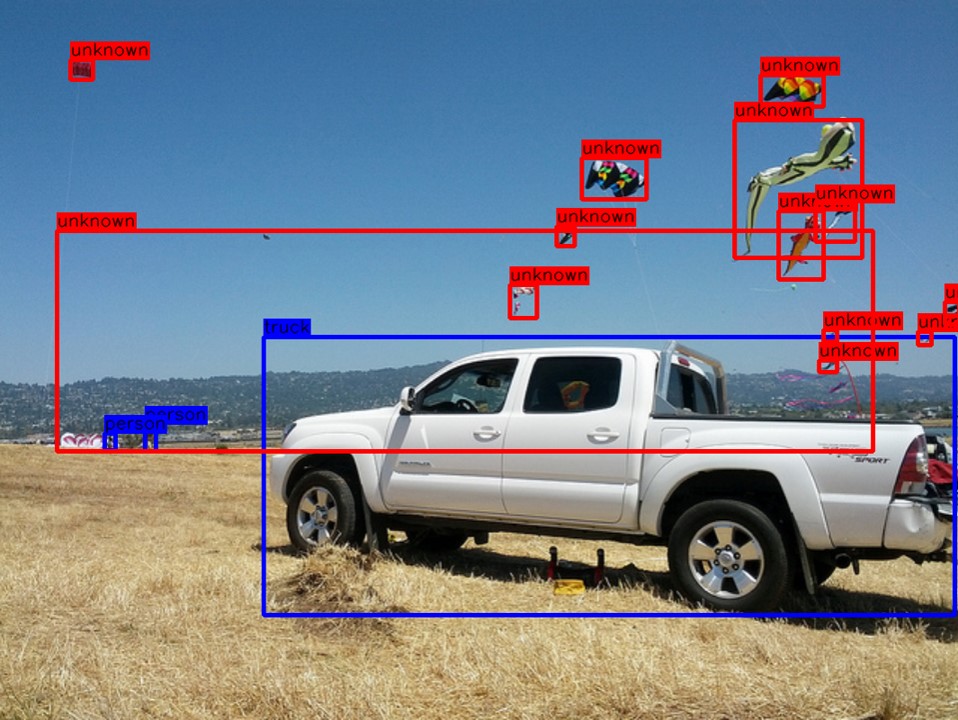}
        \subcaption{Ground Truth}
        \label{fig:ours_gt2}
    \end{minipage}
    \begin{minipage}[t]{.32\hsize}
        \centering
        \includegraphics[width=\linewidth]{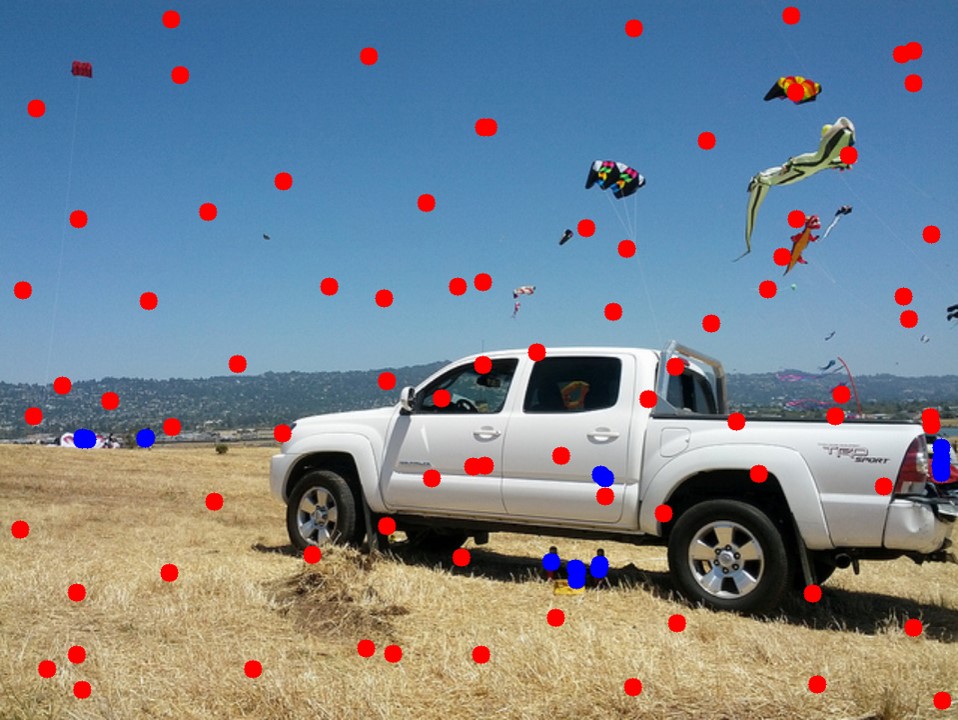}
        \subcaption{Reference points of the initial layer}
        \label{fig:ours_initref2}
    \end{minipage}
    \begin{minipage}[t]{.32\hsize}
      \centering
      \includegraphics[width=\linewidth]{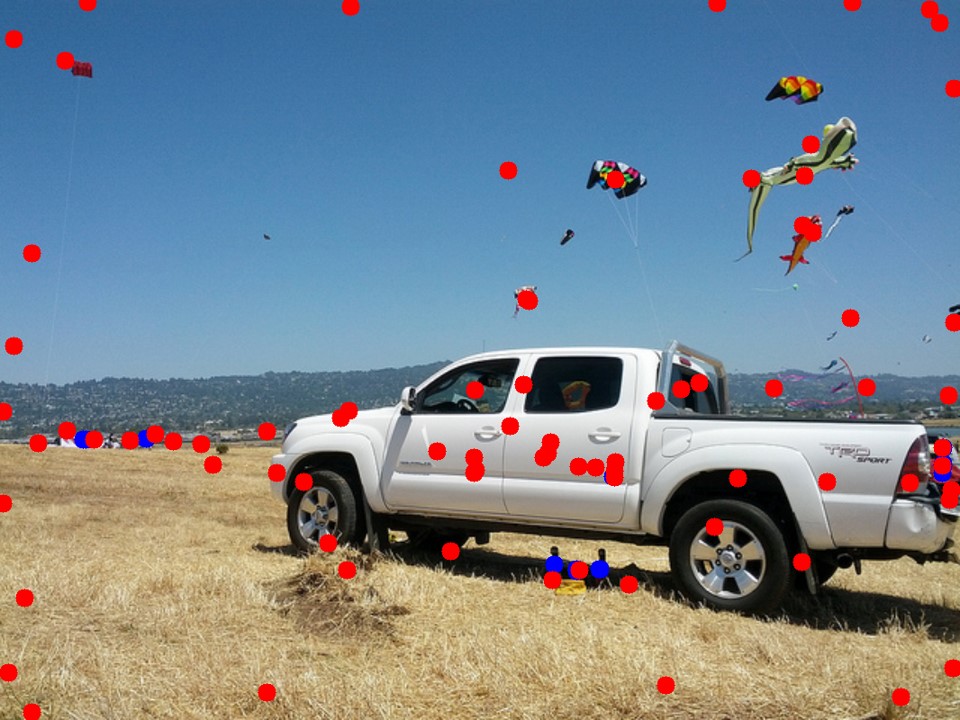}
      \subcaption{Reference points of the last layer}
      \label{fig:lastref2}
    \end{minipage} 
    \caption{\textbf{Visualization of reference points in the proposed model.}
    (a) is the ground truth image, where blue bounding boxes indicate known objects and red bounding boxes indicate unknown objects. 
    (b) and (c) show the reference points in the initial and last layers of the proposed model, respectively. 
    Blue reference points are initialized by query selection, while red reference points are initialized by learnable queries. 
    In the initial layer, reference points from query selection are distributed around known objects, and those from learnable queries are evenly distributed across the entire image. 
    In the last layer, reference points from learnable queries have moved to capture unknown objects.}
    \label{fig:first_ref_points}
\end{figure}

In recent years, object detectors have achieved rapid performance improvements alongside the development of DETR-based models\cite{carion2020end,zhao2023detrs,zhang2022dino,pu2024rank,roh2021sparse,li2023lite,chen2023group}. 
These models have enhanced detection accuracy and inference speed, making them promising for applications such as autonomous driving and robotics. 
However, in real-world scenarios, object detectors may encounter unknown objects that are not part of their training. 
It is particularly crucial for applications like autonomous driving\cite{singh2021order,ma2022rethinking} to handle unknown objects and ensure safety. 
Traditional closed-set object detectors struggle to handle such unknown objects.

To address this challenge, tasks such as Unknown Object Detection\cite{liang2023unknown,du2022unknown,kumar2023normalizing,wu2023discriminating} and Open World Object Detection (OWOD)\cite{joseph2021towards,gupta2022ow,zhao2023revisiting,ma2023cat,saito2022learning} have been proposed. 
OWOD, in particular, involves detecting both known and unknown objects and updating the model with new labels (previously unknown classes) without retraining from scratch. 
This incremental learning approach, which improves performance without altering the model's structure, is beneficial for real-world applications like autonomous driving. 
OWOD is a highly challenging task that reflects the complexities of real-world deployment while considering practical applications.

In the field of OWOD, the method of learning for unlabeled unknown objects is crucial. 
Most research has focused on training models using pseudo ground truth (GT) for unknown objects.
A pioneering study, ORE\cite{joseph2021towards}, uses a class-agnostic Region Proposal Network (RPN) in a Faster R-CNN-based model\cite{ren2016faster}, utilizing proposals with high objectness scores that do not overlap with known object GTs as pseudo unknown GTs for training.
OW-DETR\cite{gupta2022ow} pseudo-labels regions with high activation in the feature maps of the backbone, which do not match known object GTs, as unknown objects.
On the other hand, more recent methods such as PROB\cite{zohar2023prob} have been proposed for OWOD without requiring pseudo-labeling of unknown objects. 
This method learns probabilistic objectness, calculating the likelihood of each query being an object, thereby directly detecting not only known objects but also unknown objects.
Although PROB is a groundbreaking method that does not require pseudo-labeling of unknown objects, it has been pointed out that there are conflicts in learning probabilistic objectness in the deformable transformer decoder \cite{he2023usd}. 
USD\cite{he2023usd}, a method that addresses this issue, proposes a structure where objectness is predicted only in the initial layers of the decoder, and only class and bounding box are predicted in the subsequent layers. 
However, it is believed that in the initial layers of the decoder, the queries do not capture sufficient object information \cite{yao2021efficient}, making reliable objectness estimation difficult.
Therefore, this study focuses on the query selection method, which has been extensively studied in traditional object detection. 
Query selection applies class and bounding box heads to the features of the transformer encoder and uses high-class score regions to initialize object queries. 
In turn, it becomes possible for object queries to efficiently extract features from the encoder feature maps in the initial layers of the decoder.

However, in the context of OWOD, where unknown object detection is necessary, simply initializing object queries based on class scores, which focuse only on known objects, is not suitable. 
Hence, we propose Task-Decoupled Query Initialization (TDQI), which combines query selection and learnable queries to improve both known and unknown object detection performance.
As shown in \cref{fig:first_ref_points}, TDQI divides roles such that object queries initialized by query selection detect known objects, and those initialized by learnable queries cover missed known objects and unknown objects. 
By introducing TDQI, we can efficiently extract features around objects from the initial layers of the decoder, 
mitigating the issues of conducting objectness prediction only in the initial layers as done in USD. 
Additionally, to prevent conflicts between objectness and class learning, we propose Early Termination of Objectness Prediction (ETOP). 
ETOP stops the estimation of objectness in the shallow layers of the decoder. 
Unlike USD, our approach continues to predict class and bounding box in all decoder layers.
For details, refer to \cref{sec:method}.

Finally, in this study, we propose Decoupled PROB, a model based on PROB that incorporates TDQI, a novel object query initialization method for OWOD, and ETOP, a method to mitigate the learning conflict between objectness and class. Extensive experiments demonstrate that Decoupled PROB achieves state-of-the-art results on several metrics.

Our contributions are summarized as follows:

\begin{quote}
    \begin{itemize}[noitemsep,leftmargin=0pt,rightmargin=0pt]
\item By introducing TDQI, we enable efficient feature extraction around objects from the initial layers of the decoder and allow for objectness estimation in the shallow layers. 
TDQI is a module that can be easily integrated into Deformable DETR\cite{zhu2020deformable}-based OWOD models.
\item By implementing ETOP in PROB, we mitigate the learning conflicts between objectness and class predictions.
\item We propose Decoupled PROB, which incorporates both TDQI and ETOP into the PROB-based model, achieving state-of-the-art performance on several metrics in OWOD benchmarks. 
\end{itemize}
\end{quote}

\begin{figure*}[tp]
    \centering
    \begin{minipage}[t]{.18\hsize}
        \centering
        \includegraphics[width=\linewidth]{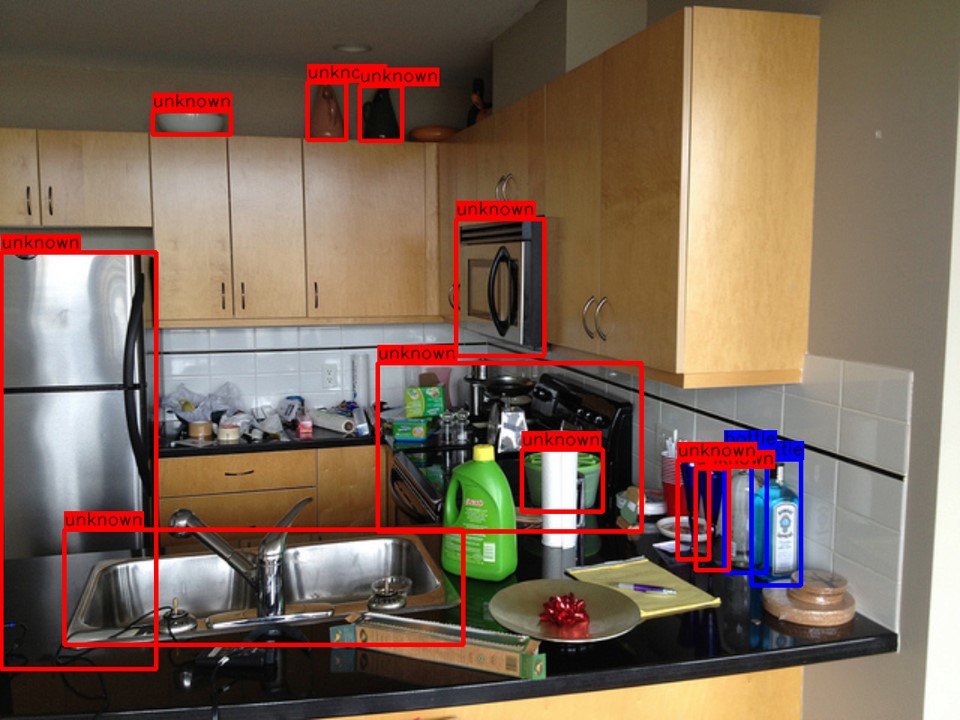}
        \label{fig:gt}
    \end{minipage}
    \centering
    \begin{minipage}[t]{.18\hsize}
        \centering
        \includegraphics[width=\linewidth]{figs/lq_qs_lqinitref1.png}
        \label{fig:refp1}
    \end{minipage}
    \begin{minipage}[t]{.18\hsize}
        \centering
        \includegraphics[width=\linewidth]{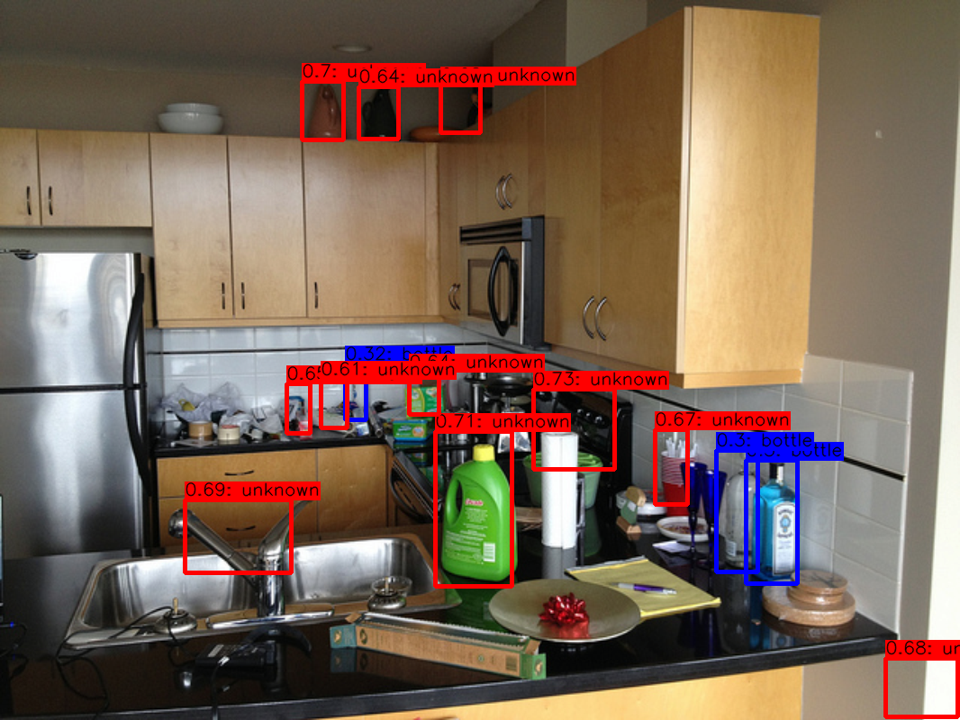}
        \label{fig:pred1}
    \end{minipage}
    \centering
    \begin{minipage}[t]{.18\hsize}
        \centering
        \includegraphics[width=\linewidth]{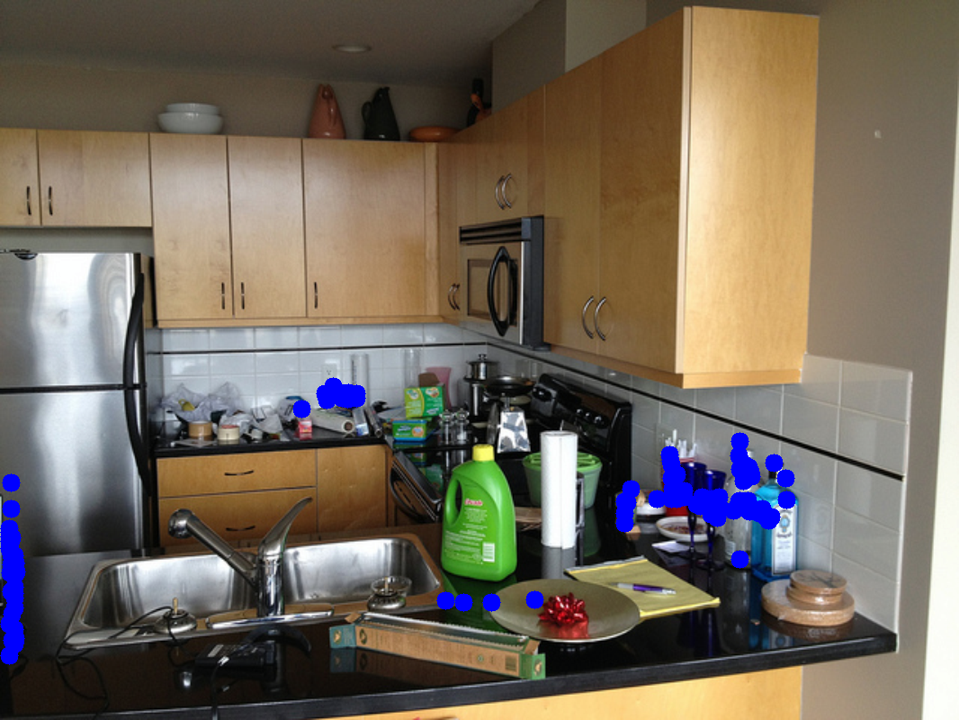}
        \label{fig:refp2}
    \end{minipage}
    \begin{minipage}[t]{.18\hsize}
        \centering
        \includegraphics[width=\linewidth]{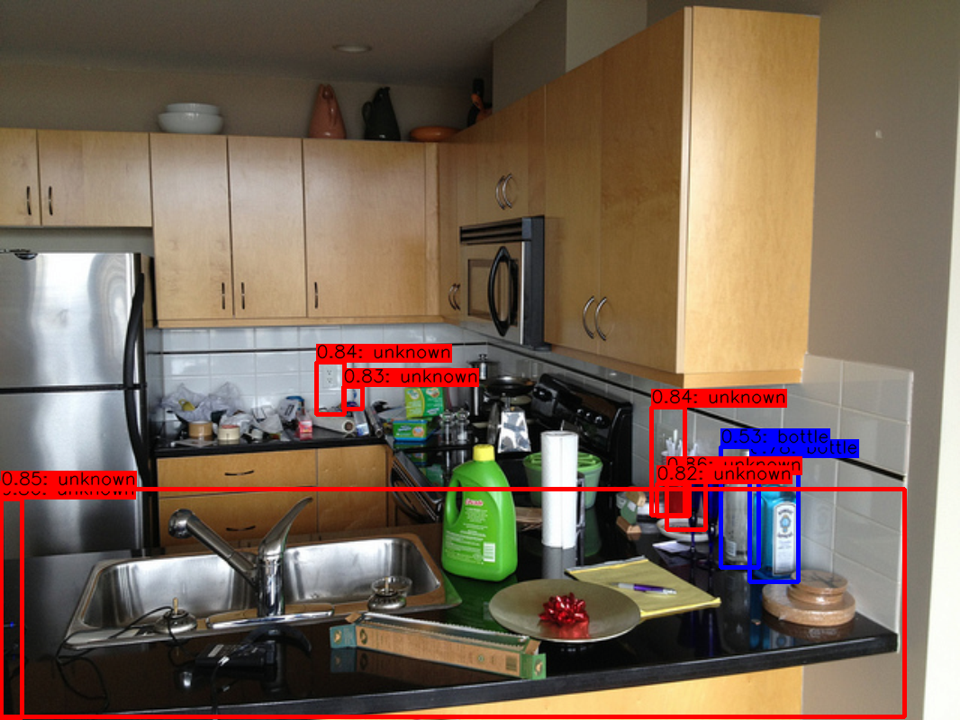}
        \label{fig:pred2}
    \end{minipage} \\

    \centering
    \begin{minipage}[t]{.18\hsize}
        \centering
        \includegraphics[width=\linewidth]{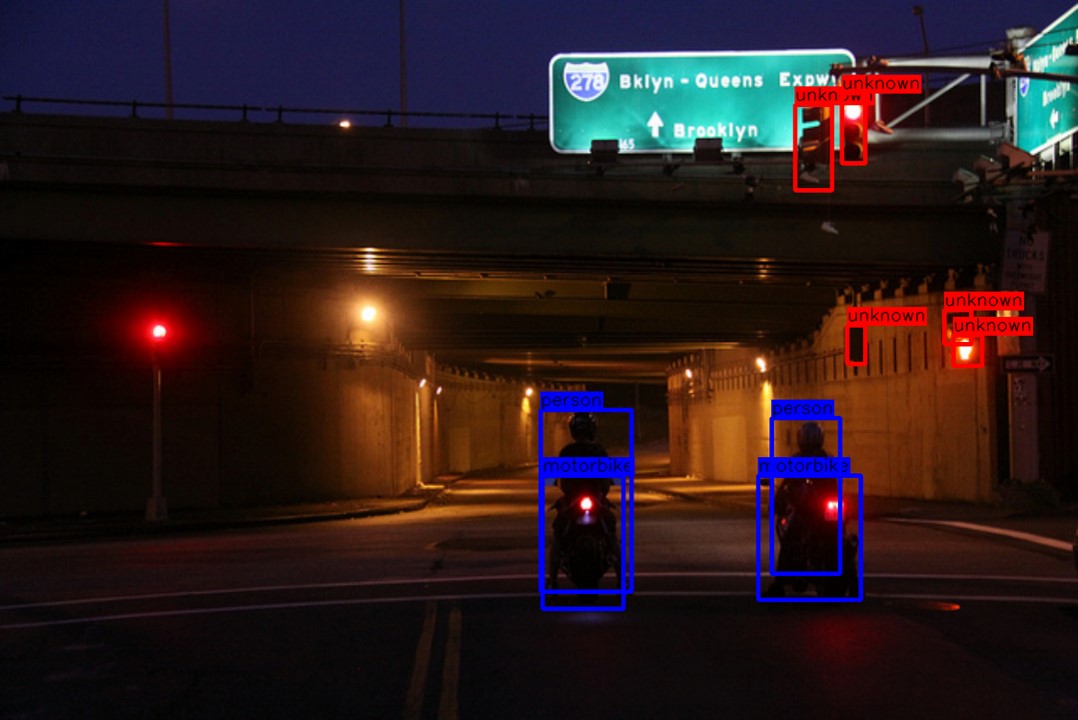}
        \subcaption{Ground truth}
        \label{fig:gt}
    \end{minipage}
    \centering
    \begin{minipage}[t]{.18\hsize}
        \centering
        \includegraphics[width=\linewidth]{figs/lq_qs_lqinitref2.png}
        \subcaption{Reference points of the learnable query model}
        \label{fig:refp1}
    \end{minipage}
    \begin{minipage}[t]{.18\hsize}
        \centering
        \includegraphics[width=\linewidth]{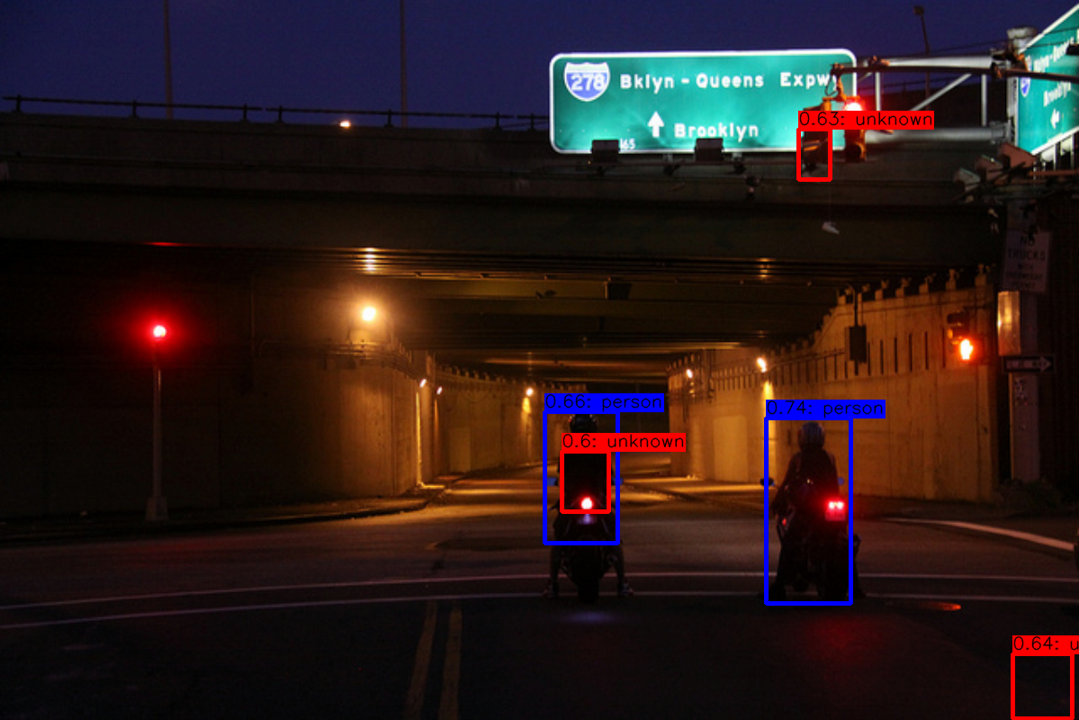}
        \subcaption{Detection results of the learnable query model}
        \label{fig:pred1}
    \end{minipage}
    \centering
    \begin{minipage}[t]{.18\hsize}
        \centering
        \includegraphics[width=\linewidth]{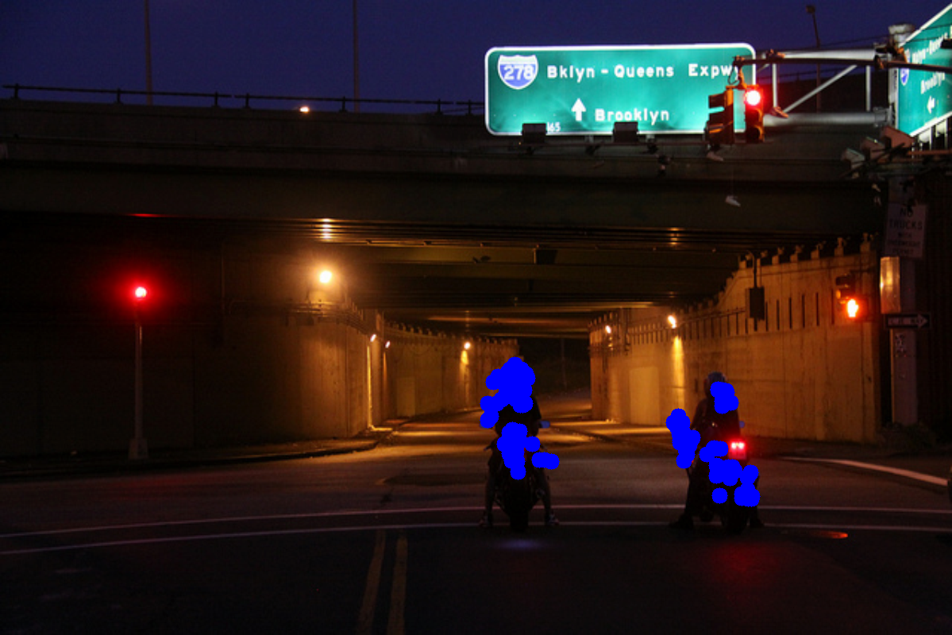}
        \subcaption{Reference points of the query selection model}
        \label{fig:refp2}
    \end{minipage}
    \begin{minipage}[t]{.18\hsize}
        \centering
        \includegraphics[width=\linewidth]{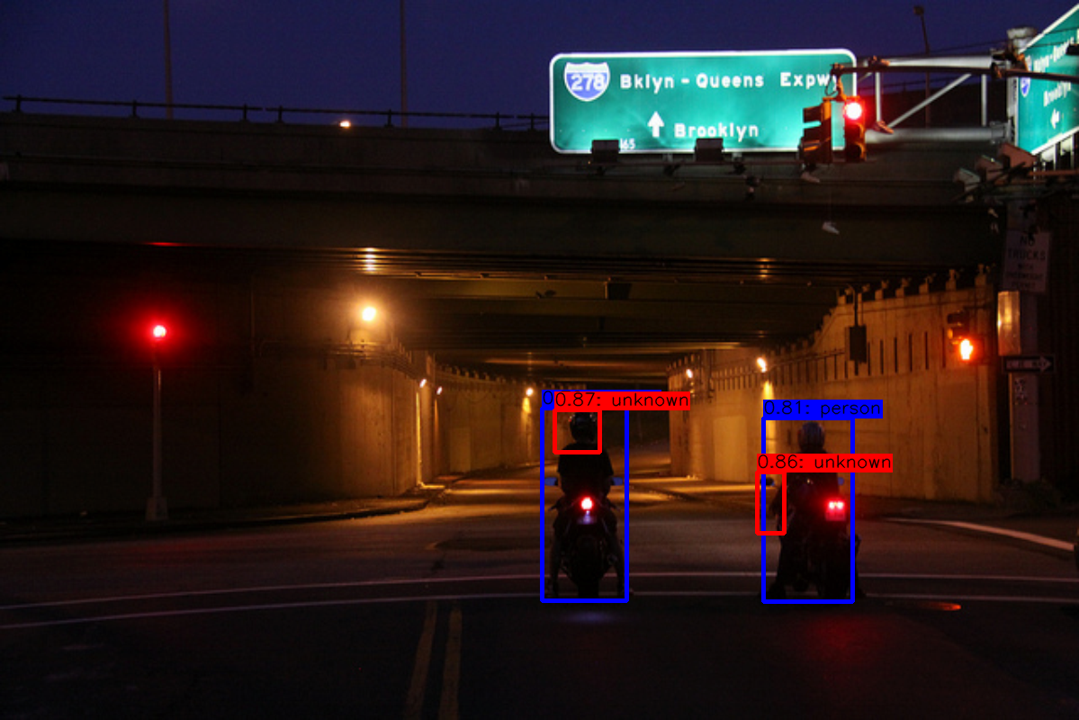}
        \subcaption{Detection results of the query selection model}
        \label{fig:pred2}
    \end{minipage}

    \caption{\textbf{Visualization of reference points and detection results for learnable query and query selection models.}
    In (a), (c), (e), blue bounding boxes indicate known objects, and red bounding boxes indicate unknown objects. 
    In the model using object queries initialized by query selection, the reference points are distributed around known objects and do not cover unknown objects, making it difficult to detect unknown objects. 
    In contrast, in the model using object queries initialized by learnable queries, the reference points are evenly distributed across the entire image, covering unknown objects as well.}
    \label{fig:lq_qs_ref}
\end{figure*}

\label{sec:intro}

\section{Related Work}

\subsection{Open World Object Detection (OWOD)}
The first model proposed along with the definition of Open World Object Detection is ORE\cite{joseph2021towards}.
ORE adapts a Faster R-CNN-based model\cite{ren2016faster} for OWOD by integrating pseudo-GT auto-labeling for unknown objects, Contrastive Clustering, and an Energy-Based Unknown Identifier (EBUI). 
Furthermore, several extensions of Faster R-CNN-based models have been researched for OWOD.
OCPL\cite{yu2022open}, proposes a prototype branch with  Proposal Embedding Aggregator and Embedding Space Compressor for clustering known classes, inspired by research in open-set recognition\cite{chen2020learning,yang2020convolutional}.
UC-OWOD\cite{wu2022uc} introduces such as Unknown Label-aware Proposal and Unknown-discriminative Classification Head for solving the problem of discovering unknown classes from the background.
2B-OCD\cite{wu2022two}, presents a localization-based objectness detection head for more accurate estimation of objectness.

In current closed-world object detectors, Transformer-based models are predominant. 
Similarly, in OWOD, models such as OW-DETR\cite{gupta2022ow} are being proposed and are becoming mainstream. 
OW-DETR, based on Deformable DETR\cite{zhu2020deformable}, introduces pseudo-GT labeling for unknown objects by leveraging the feature maps from the backbone. 
Other Transformer-based approaches include CAT\cite{ma2023cat}, which proposes a Cascade Decoupled Decoding Way to sequentially decode the localization of foreground objects and the classification of objects, including those unknown, using a shared decoder. 
CAT also introduces a pseudo-labeling method using selective search\cite{uijlings2013selective}.
Recently, PROB\cite{zohar2023prob} has been proposed, which utilizes probabilistic objectness for class-agnostic object detection, thereby eliminating the need for pseudo-labeling. 
Based on PROB, USD\cite{he2023usd} introduces Segment Anything Model (SAM)\cite{kirillov2023segment} for auxiliary labeling of unknown objects and Decoupled Objectness Learning (DOL).
DOL addresses the learning conflicts between objectness and class predictions by performing objectness predictions only in the initial layers of the decoder, while class and bounding box predictions are performed in the subsequent layers.
However, as shown in \cref{fig:lq_qs_ref}, the initial layers of the decoder which use learnable queries do not effectively extract features from object regions, indicating the insufficiency of DOL for objectness. 
Therefore, we introduce two new modules to PROB: Task-Decoupled Query Initialization (TDQI) and Early Termination of Objectness Prediction (ETOP).
TDQI allows for efficient and effective feature extraction from object regions in the initial layers of the decoder. 
Unlike USD, ETOP stops objectness estimation in the shallow layers of the decoder while estimating class and bounding box in all decoder layers, allowing for iterative refinement\cite{zhu2020deformable} and further performance improvement.

\subsection{DETR-based Object Detection}
In recent years, numerous studies\cite{liu2021dab,meng2021conditional,wang2022anchor,pu2024rank,dai2021dynamic} have been conducted in the field of closed-set object detection based on DETR\cite{carion2020end}. 
DETR is an end-to-end object detection model based on vision transformer\cite{dosovitskiy2020image}. 
Its major advantage lies in eliminating the need for handcrafted components such as non-maximum suppression, which has attracted significant attention from researchers.
However, DETR has several issues, including high computational cost, poor performance in small object detection, and slow training convergence. 
Subsequent research has primarily focused on improving these problems.
Deformable DETR\cite{zhu2020deformable} introduces a multi-scale deformable attention module, which mitigates the issues of detecting small objects and slow convergence present in DETR.
In deformable attention, reference points are generated for each object query. 
It attends only to a small set of keys called sampling points which are located around these reference points, thereby reducing the computational complexity of the attention module.
To further improve performance, Deformable DETR proposes query selection. 
In query selection, the class head and bounding box head are applied to the features obtained from the encoder. 
The features with the \textit{top-k} class scores are recorded as object queries, and the coordinates of the corresponding bounding boxes are used as reference points.
Efficient DETR\cite{yao2021efficient} achieves competitive results with fewer encoder and decoder layers compared to state-of-the-art methods by adopting query selection. 
This approach is based on a qualitative analysis showing that in the initial layers of the decoder, learnable queries distribute reference points evenly across the entire image, 
whereas query selection focuses reference points in the foreground.
Recently, high-performance object detectors such as DINO\cite{zhang2022dino} and RT-DETR\cite{zhao2023detrs} have made significant advancements, 
including new proposals related to query selection. 
DINO introduces contrastive denoising training, an evolution of the denoising training method proposed in DN-DETR\cite{li2022dn}, and implements mixed query selection, 
where positional queries are initialized using query selection and content queries are initialized as learnable queries. 
RT-DETR modifies the encoder structure to increase its efficiency one and incorporates uncertainty-minimal query selection, 
which considers uncertainty, to achieve real-time performance and high-accuracy detection.

As discussed above, query selection for object detectors has been extensively studied in the context of closed-set object detection. 
In this work, we rethink query selection from the perspective of OWOD and propose a new query initialization method adapted for the open world setting, called Task-Decoupled Query Initialization.

\label{sec:related}

\section{Preliminaries}
\noindent\textbf{Problem Formulation.} 
In the dataset $\mathcal{D}^t = \{\mathcal{I}^t, \mathcal{Y}^t\}$ given at any time $t$, the input image set $\mathcal{I}^t = \{\textbf{I}_1, ..., \textbf{I}_N\}$ corresponds to labels $\mathcal{Y}^t = \{\textbf{Y}_1, ..., \textbf{Y}_N\}$, 
each containing $\mathcal{K}$ object instance labels $\textbf{Y}_i = \{y_1, ..., y_k\}$. 
Each object instance is represented by a class label and its bounding box's center coordinates, width, and height as $y_k = [c_k, x_k, y_k, w_k, h_k]$.  
The class labels are member of the set of $C$ known object classes, $\mathcal{K}^t = \{1, 2, ..., C\}$, and during inference, the model may encounter unknown classes $\mathcal{U}^t = \{C+1, ...\}$.  

As described in \cite{joseph2021towards}, in the OWOD setting, the model $\mathcal{M}^t$ at time $t$ is trained to detect all $C$ object classes and to identify unseen class instances as unknown. 
The set of unknown instances $\mathcal{U}^t$ is labeled by an oracle with several classes of interest, creating a new training set.  
The model incrementally learns the new classes, and $\mathcal{M}^{t+1}$ is updated to detect classes $\mathcal{K}^{t+1} = \mathcal{K}^t + \{C+1, ...\}$. 
This cycle is repeated, allowing the model to update for new class detection without forgetting previously learned knowledge.

\noindent\textbf{PROB\cite{zohar2023prob}.} 
In OWOD, due to the lack of explicit supervision for unknown objects, most models need to identify these objects and use them as pseudo GT for learning.
On the other hand, PROB eliminates the need to identify unknown objects during training and removes the use of pseudo unknown object GTs by separately learning probabilistic objectness and object class.
Objectness probability $p(o|q)$ and object class probability $p(c|o,q)$ are learned separately, and the final class prediction is computed as follows:
\begin{equation}
    p(c|q) = p(c|o,q) \cdot p(o|q)
\end{equation}

The objectness probability is modeled as a multivariate gaussian distribution in the query embedding space and is calculated using the mahalanobis distance:
\begin{equation}
    \begin{split}
    f_{obj}(\bm{q}) &= \exp\left(- (\bm{q} - \bm{\mu})^T \bm{\Sigma}^{-1} (\bm{q} - \bm{\mu}) \right) \\
        &= \exp(d_M(\bm{q})^2) 
    \end{split}
\end{equation}
where $\bm{\mu}$ is the mean vector and $\bm{\Sigma}$ is the covariance matrix of the query embeddings, and $d_M(\bm{q})$ is the mahalanobis distance of the query embedding.
Training is conducted in an alternating two-step process: \textbf{1.} Estimating the mean $\bm{\mu}$ and covariance $\bm{\Sigma}$ of the query embeddings.
\textbf{2.} Maximizing the likelihood of matched query embeddings by penalizing the squared Mahalanobis distance:
\begin{equation}
    \mathcal{L}_{obj} = \sum_{i \in Z} d_M(q_i)^2
\end{equation}
where $\mathcal{L}_{obj}$is the objectness loss, and $Z$ is the set of query embeddings matched with the GT.

\begin{figure}[tp]
    \centering
    \includegraphics[width=0.97\linewidth]{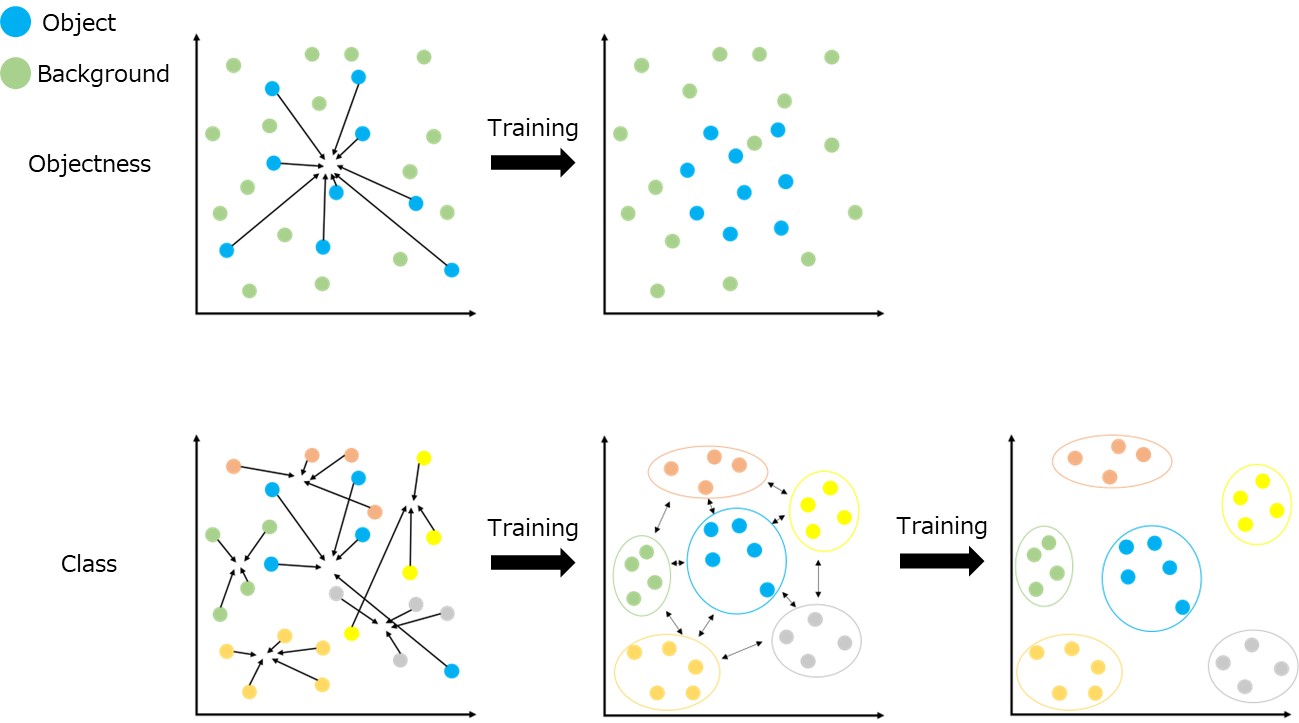}
    \caption{\textbf{Learning of objectness and class classification.} 
    Objectness aims to separate objects from the background by bringing the features of objects closer together, 
    whereas class classification aims to separate the features among different object classes. 
    Therefore, these processes work in opposite directions.}
    \label{fig:objectness_and_class}
\end{figure}

\label{sec:background}

\section{Method}

\begin{figure*}[tp]
    \centering
    \includegraphics[width=0.97\linewidth]{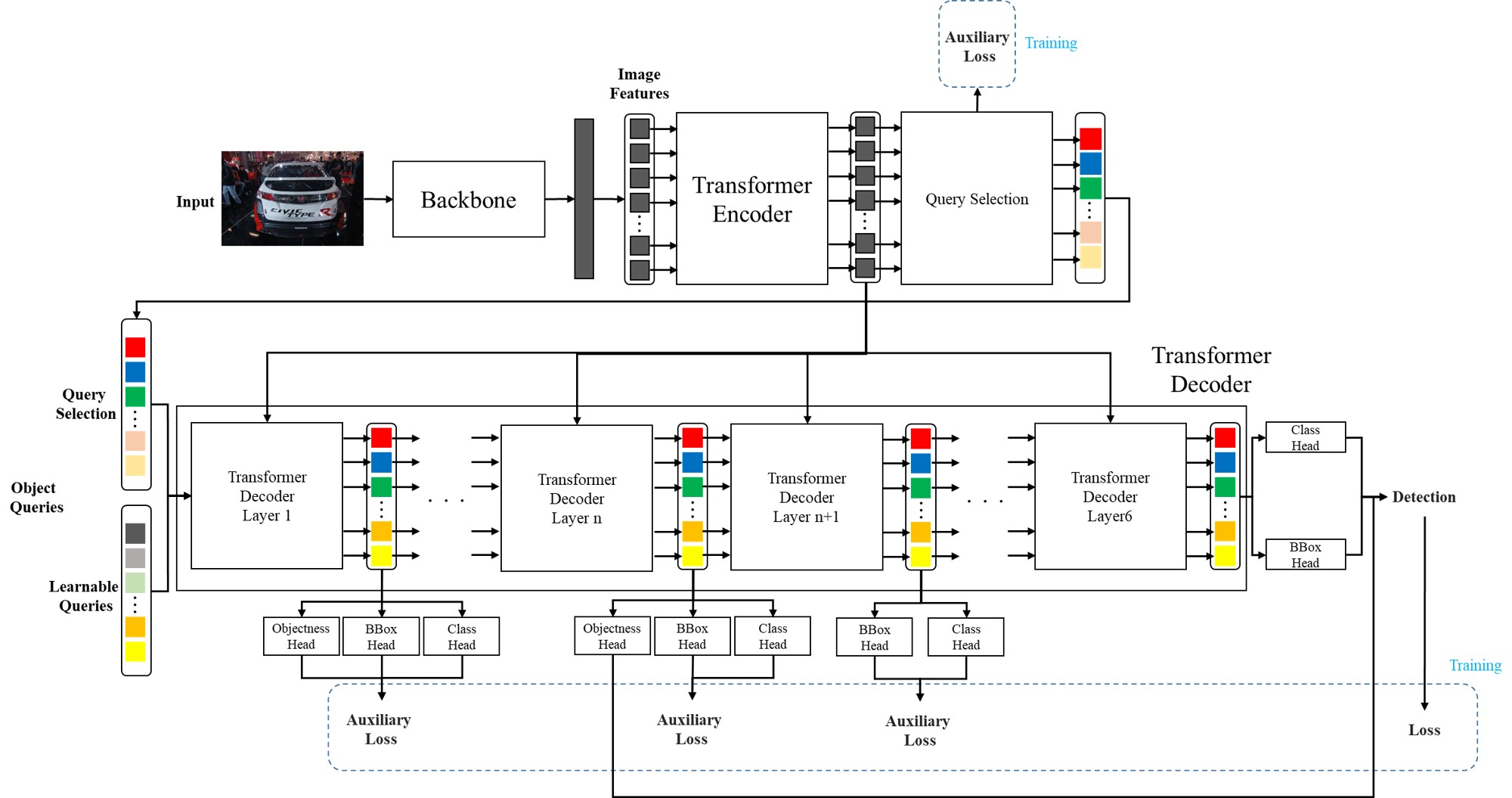}
    \caption{\textbf{Overview of our proposed model.} 
    Our model processes the input image with a backbone and a deformable transformer encoder, same as PROB. 
    Then, query selection is performed on the encoder features, which are concatenated with learnable queries to form the object queries. 
    In the deformable transformer decoder, ``Layer $n$'' represents the layer where objectness prediction stops, set to $n=2$ based on experimental results (\cref{tab:num_obj}). 
    Thus, in the decoder, the first two layers predict objectness, class, and bounding box, while the subsequent layers predict only class and bounding box. 
    The objectness predicted in the second layer is used for both training and inference, along with the class and bounding box predictions from the last layer.}
    \label{fig:model}
\end{figure*}

\begin{table}[tp]
    \centering
    \caption{\textbf{Results of simply initializing PROB's object queries with query selection.} 
    We compare a model where PROB's object queries are simply initialized with query selection on the M-OWODB task 1 (explained in \cref{sec:experiment}). 
    In the OWOD task, simply introducing query selection based on class scores causes a decrease in unknown object detection performance.}
    \begin{tabular}{ccc}
        \toprule
        \textbf{Methods} & \textbf{U-Recall(↑)} & \textbf{mAP(↑)} \\
        \midrule
        PROB & 19.4 & 59.5 \\
        PROB+Query Selection & 14.5 & 59.9 \\
        \bottomrule
    \end{tabular}
    \label{tab:prob_qs}
\end{table}

We propose Decoupled PROB, which is based on PROB and incorporates TDQI and ETOP. 
In this section, we describe ETOP, TDQI, and the model architecture. 
Details for model training are described in the supplementary material.

\noindent \textbf{Early Termination of Objectness Prediction (ETOP).} 
PROB refines the information held by each object query embeddings by predicting bounding box, class, and objectness at each decoder layer and learning based on auxiliary losses. 
However, as suggested in \cite{he2023usd}, we believe that simultaneously predicting class and objectness in \underline{all decoder layers} can cause conflicts between these predictions.
\cref{fig:objectness_and_class} illustrates the relationship between class and objectness. 
Class prediction learns to separate features among different object classes, while objectness prediction learns to bring object features closer together.
USD\cite{he2023usd} proposes estimating only objectness in the initial decoder layer and then estimating class and bounding box in subsequent layers to avoid conflicts between class and objectness predictions.
Similarly, we propose a method to reduce conflicts between objectness and class learning by stopping objectness estimation in the shallow layers of the decoder, which we call Early Termination of Objectness Prediction (ETOP).
However, as indicated in \cite{yao2021efficient}, when using learnable queries to initialize object queries as in PROB and USD, the initial layers of the decoder do not adequately capture object features, 
meaning that objectness predictions are performed before the object features are adequately extracted. 
Therefore, as qualitatively demonstrated in \cite{yao2021efficient}, we introduce query selection to distribute reference points around the foreground from the initial layers of the decoder, 
allowing the objectness prediction to accurately capture object features from an early stage.
Since query selection requires the prediction and iterative refinement of class scores and bounding box predictions, 
unlike USD, ETOP continues to predict class and bounding box in the layers where objectness is predicted, and these predictions are used in training.

\noindent \textbf{Task-Decoupled Query Initialization (TDQI).} 
We followed a simple approach similar to \cite{zhu2020deformable} and \cite{yao2021efficient}, 
where queries are initialized based on \textit{top-k} selection using class scores, to train and evaluate the OWOD task. 
However, as shown in \cref{tab:prob_qs}, merely applying query selection to the OWOD model does not improve performance. 
As illustrated in \cref{fig:lq_qs_ref}, the reason is that reference points initialized by query selection are distributed around known objects since they are based on class scores.
This prevents object query embeddings from capturing features around unknown objects.
In contrast, when using learnable queries, reference points are evenly distributed across the entire image, 
allowing the model to capture features of unknown objects that could be located anywhere in the image. 
Therefore, we propose Task-Decoupled Query Initialization (TDQI) that combines query selection, which places reference points around objects from the early stages and improves object detection performance, 
with learnable queries, which distribute reference points evenly across the image and cover unknown objects.
The object queries initialized by query selection primarily detect known objects, while those initialized by learnable queries cover missed known objects as well as unknown objects. 
This combination aims to further improve both known and unknown object detection performance.
Additionally, since the class head performs $(C+1)$ classification for $C$ known object classes plus one additional class reserved for unknown objects and background, 
the scores for the additional class are disregarded in our query selection, which is responsible for detecting known objects.

\noindent \textbf{Proposed Model Architecture.} 
Our final model structure is shown in \cref{fig:model}. 
We introduce TDQI to the base model PROB, and further stop the prediction of objectness in the shallow layers of the decoder (ETOP). 
In OWOD models based on Deformable DETR such as PROB, the number of object queries is standardized to 100. 
We adhere to this by setting the total number of object queries initialized by query selection and learnable queries to 100.
Additionally, for fairness, the number of decoder layers is set to 6, as in \cite{gupta2022ow,zohar2023prob,he2023usd}.

The number of object queries initialized by query selection and learnable queries, and the layer at which objectness prediction is stopped, 
are determined based on the quantitative experimental results shown in \cref{tab:qs_lq} and \cref{tab:num_obj}.
We simply select the structure that yields the highest combined U-Recall and mAP. 
As shown in \cref{tab:qs_lq} and \cref{tab:num_obj}, the best configuration is to initialize 20 queries via query selection, 80 via learnable queries, 
to stop objectness estimation after 2 layers. 
This configuration is used in our final proposed model.

\begin{table}[tp]
    \centering
    \caption{\textbf{Comparison of the ratio of Query Selection to Learnable Queries.} 
    Experimental results on the ratio of query selection to learnable queries in Task 1 of M-OWODB (explained in \cref{sec:experiment}) are shown. 
    The total number of object queries is set to 100, consistent with \cite{zohar2023prob,ma2023cat,gupta2022ow}. 
    In these experiments, the objectness stop layer for ETOP in Decoupled PROB is set to 2.}
    \begin{adjustbox}{width=1.\linewidth,center}
    \begin{tabular}{cc|cc}
        \toprule
        \textbf{Query Selection} & \textbf{Learnable Query} & \textbf{U-Recall(↑)} & \textbf{mAP(↑)} \\
        \midrule
        10 & 90 & \textbf{20.4} & 58.8 \\
        20 & 80 & 20.3 & 59.8 \\
        30 & 70 & 19.4 & \textbf{59.9} \\
        40 & 60 & 18.8 & \textbf{59.9} \\
        50 & 50 & 18.8 & 59.7 \\
        70 & 30 & 18.8 & 59.7 \\
        90 & 10 & 18.5 & 59.5 \\
        \bottomrule
    \end{tabular}
  \end{adjustbox}
  \label{tab:qs_lq}
  \end{table}
  
  \begin{table}[tp]
    \centering
    \caption{\textbf{Comparison of objectness stop layers in ETOP.} 
    Experimental results on the objectness stop layer for ETOP are shown. 
    A comparison is made using Task 1 of M-OWODB (explained in \cref{sec:experiment}). 
    In this experiment, the query selection and learnable queries are set to 20 and 80, respectively.}
    \scalebox{0.9}[0.8]{
    \begin{tabular}{c|cc}
        \toprule
        \textbf{Layer Number} & \textbf{U-Recall(↑)} & \textbf{mAP(↑)} \\
        \midrule
        1 & \textbf{20.7} & 58.0 \\
        2 & 20.3 & 59.8 \\
        3 & 19.0 & 59.7 \\
        4 & 18.9 & 60.1 \\
        5 & 17.9 & 59.8 \\
        6 & 18.8 & \textbf{60.4} \\
        \bottomrule
    \end{tabular}
    }
  \label{tab:num_obj}
\end{table}


\label{sec:method}

\section{Experiments and Results} \label{sec:Experiments}

\begin{table*}[t]
  \centering
  \caption{\textbf{State-of-the-Art comparison for OWOD on M-OWODB (top) and S-OWODB (bottom).}
  The comparison is shown in terms of mAP@0.5 for known classes and recall (U-Recall) for unknown classes. 
  The known classes metrics include previously, currently, and all known objects. 
  Note that U-Recall is not calculated for Task 4 as all 80 classes are considered known. 
  \textbf{Bold text} in the table indicates the best performance for each metric, while \underline{underlined text} represents the second-best performance.
  Decoupled PROB demonstrates state-of-the-art results in many metrics, significantly outperforming other models in known object detection performance.}
  \begin{adjustbox}{width=0.97\linewidth,center}
    \begin{tabular}{
      l|
      >{\columncolor{yellow!20}}c
      c
      c|
      >{\columncolor{yellow!20}}c
      c
      c
      c|
      >{\columncolor{yellow!20}}c
      c
      c
      c|
      c
      c
      c}
      \toprule
      \textbf{Task IDs (→)} & \multicolumn{3}{c|}{\textbf{Task 1}} & \multicolumn{4}{c|}{\textbf{Task 2}} & \multicolumn{4}{c|}{\textbf{Task 3}} & \multicolumn{3}{c}{\textbf{Task 4}} \\
      \midrule
      & U-Recall & \multicolumn{2}{c|}{\cellcolor{cyan!10}mAP (↑)} & U-Recall & \multicolumn{3}{c|}{\cellcolor{cyan!10}mAP (↑)} & U-Recall & \multicolumn{3}{c|}{\cellcolor{cyan!10}mAP (↑)} & \multicolumn{3}{c}{\cellcolor{cyan!10}mAP (↑)} \\
      & (↑) & \multicolumn{2}{c|}{\shortstack{Current \\ known}} & (↑) & \shortstack{Previously \\ known} & \shortstack{Current \\ known} & Both & (↑) & \shortstack{Previously \\ known} & \shortstack{Current \\ known} & Both & \shortstack{Previously \\ known} & \shortstack{Current \\ known} & Both \\
      \midrule
      ORE\texttt{-EBUI} \cite{joseph2021towards} & 4.9 & \multicolumn{2}{c|}{56.0} & 2.9 & 52.7 & 26.0 & 39.4 & 3.9 & 38.2 & 12.7 & 29.7 & 29.6 & 12.4 & 25.3 \\
      UC-OWOD \cite{wu2022uc} & 2.4 & \multicolumn{2}{c|}{50.7} & 3.4 & 33.1 & 30.5 & 31.8 & 8.7 & 28.8 & 16.3 & 24.6 & 25.6 & 15.9 & 23.2 \\
      OCPL \cite{yu2022open} & 8.26 & \multicolumn{2}{c|}{56.6} & 7.65 & 50.6 & 27.5 & 39.1 & 11.9 & 38.7 & 14.7 & 30.7 & 30.7 & 14.4 & 26.7 \\
      2B-OCD \cite{wu2022two} & 12.1 & \multicolumn{2}{c|}{56.4} & 9.4 & 51.6 & 25.3 & 38.5 & 12.6 & 37.2 & 13.2 & 29.2 & 30.0 & 13.3 & 25.8 \\
      OW-DETR \cite{gupta2022ow} & 7.5 & \multicolumn{2}{c|}{59.2} & 6.2 & 53.6 & \underline{33.5} & 42.9 & 5.7 & 38.3 & 15.8 & 30.8 & 31.4 & 17.1 & 27.8 \\
      CAT \cite{ma2023cat} & \textbf{23.7} & \multicolumn{2}{c|}{\textbf{60.0}} & \underline{19.1} & 55.5 & 32.7 & \underline{44.1} & \textbf{24.4} & 42.8 & 18.7 & 34.8 & 34.4 & 16.6 & 29.9 \\
      USD\texttt{-ASF} \cite{he2023usd} & \underline{21.6} & \multicolumn{2}{c|}{\underline{59.9}} & \textbf{19.7} & \textbf{56.6} & 32.5 & \underline{44.6} & \underline{23.5} & \underline{43.5} & 21.9 & \underline{36.3} & 35.4 & \underline{18.9} & 31.3 \\
      PROB(Baseline) \cite{zohar2023prob} & 19.4 & \multicolumn{2}{c|}{59.5} & 17.4 & 55.7 & 32.2 & 44.0 & 19.6 & 43.0 & \underline{22.2} & 36.0 & \underline{35.7} & \underline{18.9} & \underline{31.5} \\
      \textbf{Ours: Decoupled PROB} & 20.3 & \multicolumn{2}{c|}{59.8} & 18.4 & \underline{56.4} & \textbf{36.7} & \textbf{46.6} & 20.3 & \textbf{44.6} & \textbf{26.1} & \textbf{38.5} & \textbf{37.8} & \textbf{22.1} & \textbf{33.8} \\
      \midrule
      \midrule
      ORE\texttt{-EBUI}  \cite{joseph2021towards} & 1.5 & \multicolumn{2}{c|}{61.4} & 3.9 & 56.5 & 26.1 & 40.6 & 3.6 & 38.7 & 23.7 & 33.7 & 33.6 & 26.3 & 31.8 \\
      OW-DETR \cite{gupta2022ow} & 5.7 & \multicolumn{2}{c|}{71.5} & 6.2 & 62.8 & 27.5 & 43.8 & 6.9 & 45.2 & 24.9 & 38.5 & 38.2 & 28.1 & 33.1 \\
      CAT \cite{ma2023cat} & \textbf{24.0} & \multicolumn{2}{c|}{\textbf{74.2}} & \underline{23.0} & \textbf{67.6} & 35.5 & 50.7 & 24.6 & \textbf{51.2} & \underline{32.6} & \underline{45.0} & \underline{45.4} & \underline{35.1} & \underline{42.8} \\
      USD\texttt{-ASF} \cite{he2023usd} & \underline{19.2} & \multicolumn{2}{c|}{72.9} & 22.4 & 64.9 & \underline{38.9} & \underline{51.2} & \textbf{25.4} & 50.1 & 34.7 & 45.0 & 43.4 & 33.6 & 40.9 \\
      PROB(Baseline) \cite{zohar2023prob} & 17.6 & \multicolumn{2}{c|}{73.4} & 22.3 & \underline{66.3} & 36.0 & 50.4 & \underline{24.8} & 47.8 & 30.4 & 42.0 & 42.6 & 31.7 & 39.9 \\
      \textbf{Ours: Decoupled PROB} & 17.4 & \multicolumn{2}{c|}{\underline{73.7}} & \textbf{24.4} & \underline{66.3} & \textbf{45.7} & \textbf{55.5} & 23.5 & \underline{50.9} & \textbf{43.3} & \textbf{48.4} & \textbf{46.1} & \textbf{39.8} & \textbf{44.6} \\
      \bottomrule
    \end{tabular}
  \end{adjustbox}
\label{tab:sota_comp}
\end{table*}

We perform comprehensive experiments on OWOD benchmarks and incremental object detection tasks. 
The comparison results of our method with existing approaches are presented. 
Ablation studies and the comparison between ETOP and DOL are included in the supplementary material.

\subsection{Datasets and Metrics}
\noindent \textbf{Datasets.} 
We evaluate our proposed model using two benchmarks: the ``superclass-mixed OWOD benchmark'' (M-OWODB) and the ``superclass-separated OWOD benchmark'' (S-OWODB). 
The M-OWODB consists of images from MS-COCO\cite{lin2014microsoft} and Pascal VOC 2007\cite{everingham2010pascal}, which are divided into four non-overlapping task sets \(\{ T_1, ..., T_4\}\). 
Classes for task \( T_t \) are not introduced until task \( t \) is reached.
In each task \( T_t \), an additional 20 classes are introduced. During training for task \( t \), only these new classes are labeled. 
However, in the test set, all classes encountered in previous tasks \(\{ T_\lambda : \lambda \leq t \}\) must be detected.

In S-OWODB, object categories with closely related semantics are grouped into tasks, resulting in a clear separation of super-categories. 
To construct S-OWODB, only the MS-COCO dataset is used, and a distinct separation of super-categories is implemented. 
However, to maintain the integrity of each superclass, the number of classes introduced per increment varies.

\noindent \textbf{Evaluation Metrics.}
We use mean average precision (mAP), same as standard object detection, for known objects, and recall (U-Recall) for unknown object detection.
U-Recall evaluates the ratio of detected to total labeled unknown objects, measuring the model's ability to discover unknown object instances in the OWOD problem.

\begin{figure*}[tp]
  \centering
  \includegraphics[height=0.44\textwidth, width=1.\linewidth]{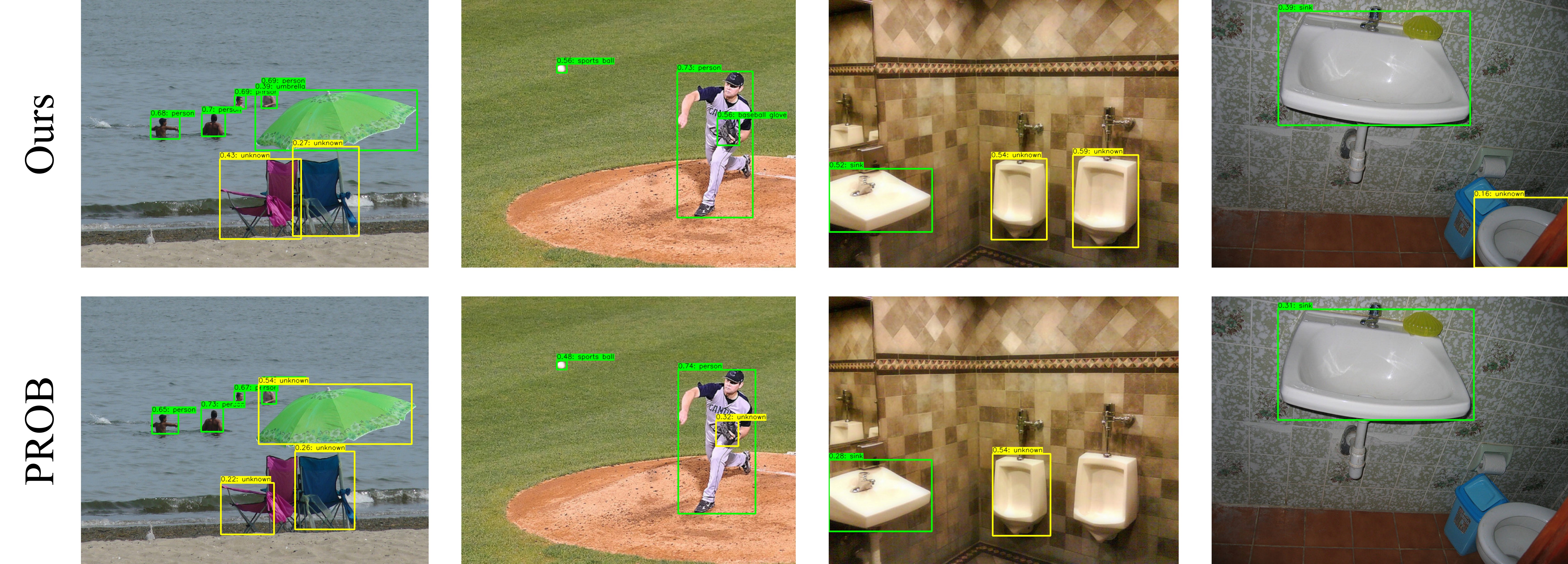}
  \caption{\textbf{Qualitative results on selected images from the test set.}
  The top row shows the detection results of Decoupled PROB, while the bottom row shows the detection results of PROB. 
  Green bounding boxes indicate known classes, and yellow bounding boxes indicate unknown classes. 
  In the left two images, Decoupled PROB accurately detects objects as known classes that PROB identifies as unknown. 
  Additionally, in the right two images, Decoupled PROB detects unknown objects that PROB fails to detect.}
  \label{fig:comp}
\end{figure*}

\subsection{Implementation Details}
Same as \cite{gupta2022ow,zohar2023prob,ma2023cat}, we use the Deformable DETR model\cite{zhu2020deformable}. 
The backbone consists of a ResNet-50\cite{he2016deep} pre-trained with DINO\cite{caron2021emerging}. The number of object queries and the query embedding dimension are set to 100 and 256, respectively. 
Like PROB, the embedding probability distribution is predicted by using the exponential moving average of the mean and covariance of the query embeddings in each mini-batch, with a momentum of 0.1.
Our model is trained with the AdamW optimizer\cite{loshchilov2018decoupled}, configured with a learning rate of $2\times10^{-3}$, $\beta_1=0.9$, $\beta_2=0.999$, and a weight decay of $10^{-4}$, same as PROB. 
Additional details are provided in the supplementary material.

\begin{table*}[tp]
  \centering
  \caption{\textbf{State-of-the-Art comparison for incremental object detection (iOD) on PASCAL VOC.}
  The comparison is presented using metrics for per-class AP and overall mAP.
  \textbf{Bold text} in the table indicates the best performance, while \underline{underlined text} represents the second-best performance.}
  \begin{adjustbox}{width=0.97\linewidth,center}
  \begin{tabular}{lccccccccccccccccccccc}
  \toprule
  \textbf{\textcolor{green}{10 + 10 setting}} & aero & cycle & bird & boat & bottle & bus & car & cat & chair & cow & \cellcolor[gray]{0.95}table & \cellcolor[gray]{0.95}dog & \cellcolor[gray]{0.95}horse & \cellcolor[gray]{0.95}bike & \cellcolor[gray]{0.95}person & \cellcolor[gray]{0.95}plant & \cellcolor[gray]{0.95}sheep & \cellcolor[gray]{0.95}sofa & \cellcolor[gray]{0.95}train & \cellcolor[gray]{0.95}tv & mAP \\
  \midrule
  ILOD \cite{shmelkov2017incremental} & 69.9 & 70.4 & 69.4 & 54.3 & 48 & 68.7 & 78.9 & 68.4 & 45.5 & 58.1 & \cellcolor[gray]{0.95}59.7 & \cellcolor[gray]{0.95}72.7 & \cellcolor[gray]{0.95}73.5 & \cellcolor[gray]{0.95}73.2 & \cellcolor[gray]{0.95}66.3 & \cellcolor[gray]{0.95}29.5 & \cellcolor[gray]{0.95}63.4 & \cellcolor[gray]{0.95}61.6 & \cellcolor[gray]{0.95}69.3 & \cellcolor[gray]{0.95}62.2 & 63.2 \\
  Faster ILOD \cite{peng2020faster} & 72.8 & 75.7 & 71.2 & 60.5 & 61.7 & 70.4 & 83.3 & 76.6 & 53.1 & 72.3 & \cellcolor[gray]{0.95}56.7 & \cellcolor[gray]{0.95}73.6 & \cellcolor[gray]{0.95}70.9 & \cellcolor[gray]{0.95}66.8 & \cellcolor[gray]{0.95}67.6 & \cellcolor[gray]{0.95}46.1 & \cellcolor[gray]{0.95}63.4 & \cellcolor[gray]{0.95}48.1 & \cellcolor[gray]{0.95}57.1 & \cellcolor[gray]{0.95}43.6 & 62.1 \\
  ORE\texttt{-(CC+EBUI)} \cite{joseph2021towards} & 53.3 & 69.2 & 62.4 & 51.8 & 52.9 & 73.6 & 83.7 & 71.7 & 42.8 & 66.8 & \cellcolor[gray]{0.95}46.8 & \cellcolor[gray]{0.95}59.9 & \cellcolor[gray]{0.95}65.5 & \cellcolor[gray]{0.95}66.1 & \cellcolor[gray]{0.95}68.6 & \cellcolor[gray]{0.95}29.8 & \cellcolor[gray]{0.95}55.1 & \cellcolor[gray]{0.95}51.6 & \cellcolor[gray]{0.95}65.3 & \cellcolor[gray]{0.95}51.5 & 59.4 \\
  ORE\texttt{-EBUI}\cite{joseph2021towards} & 63.5 & 70.9 & 58.9 & 42.9 & 34.1 & 76.2 & 80.7 & 76.3 & 34.1 & 66.1 & \cellcolor[gray]{0.95}56.1 & \cellcolor[gray]{0.95}70.4 & \cellcolor[gray]{0.95}80.2 & \cellcolor[gray]{0.95}72.3 & \cellcolor[gray]{0.95}81.8 & \cellcolor[gray]{0.95}42.7 & \cellcolor[gray]{0.95}71.6 & \cellcolor[gray]{0.95}68.1 & \cellcolor[gray]{0.95}77 & \cellcolor[gray]{0.95}67.7 & 64.5\\
  OW-DETR \cite{gupta2022ow} & 61.8 & 69.1 &67.8 & 45.8 & 47.3 & 78.3 & 78.4 & 78.6 & 36.2 & 71.5 & \cellcolor[gray]{0.95}57.5 & \cellcolor[gray]{0.95}75.3 & \cellcolor[gray]{0.95}76.2 & \cellcolor[gray]{0.95}77.4 & \cellcolor[gray]{0.95}79.5 & \cellcolor[gray]{0.95}40.1 & \cellcolor[gray]{0.95}66.8 & \cellcolor[gray]{0.95}66.3 & \cellcolor[gray]{0.95}75.6 & \cellcolor[gray]{0.95}64.1 & 65.7 \\
  PROB \cite{zohar2023prob} & 70.4 & 75.4 & 67.3 & 48.1 & 55.9 & 73.5 & 78.5 & 75.4 & 42.8 & 72.2& \cellcolor[gray]{0.95}64.2 & \cellcolor[gray]{0.95}73.8 & \cellcolor[gray]{0.95}76.0 & \cellcolor[gray]{0.95}74.8 & \cellcolor[gray]{0.95}73.5 & \cellcolor[gray]{0.95}40.2 & \cellcolor[gray]{0.95}66.2 & \cellcolor[gray]{0.95}73.3 & \cellcolor[gray]{0.95}64.4 & \cellcolor[gray]{0.95}64.0 & \underline{66.5} \\
  \midrule
  \textbf{Ours: Decoupled PROB} & 75.8 & 74.1 & 67.8 & 54.6 & 44.8 & 76.9 & 82.9 & 83.3 & 46.1 & 70.3 & \cellcolor[gray]{0.95}62.9 & \cellcolor[gray]{0.95}71.3 & \cellcolor[gray]{0.95}72.4 & \cellcolor[gray]{0.95}75.0 & \cellcolor[gray]{0.95}73.9 & \cellcolor[gray]{0.95}36.8 & \cellcolor[gray]{0.95}66.1 & \cellcolor[gray]{0.95}64.7 & \cellcolor[gray]{0.95}74.5 & \cellcolor[gray]{0.95}68.3  & \textbf{67.1} \\
  \bottomrule
  \end{tabular}
  \end{adjustbox}
  
  \vspace{0.03cm}
  
  \begin{adjustbox}{width=0.97\linewidth,center}
  \begin{tabular}{lccccccccccccccccccccc}
  \toprule
  \textbf{\textcolor{green}{15 + 5 setting}} & aero & cycle & bird & boat & bottle & bus & car & cat & chair & cow & table & dog & horse & bike & erson & \cellcolor[gray]{0.95}plant & \cellcolor[gray]{0.95}sheep & \cellcolor[gray]{0.95}sofa & \cellcolor[gray]{0.95}train & \cellcolor[gray]{0.95}tv & mAP \\
  \midrule
  ILOD \cite{shmelkov2017incremental} & 70.5 & 79.2 & 68.8 & 59.1 & 53.2 & 75.4 & 79.4 & 78.4 & 46.6 & 59.4 & 59.5 & 75.8 & 71.8 & 78.6 & 69.6 & \cellcolor[gray]{0.95}33.7 & \cellcolor[gray]{0.95}61.5 & \cellcolor[gray]{0.95}63.1 & \cellcolor[gray]{0.95}71.7 & \cellcolor[gray]{0.95}62.2 & 65.8 \\
  Faster ILOD \cite{peng2020faster} & 66.5 & 78.1 & 71.8 & 54.6 & 61.4 & 68.4 & 82.7 & 82.1 & 52.1 & 73.1 & 58.4 & 73.6 & 77.8 & 78.4 & 69.3 & \cellcolor[gray]{0.95}37.6 & \cellcolor[gray]{0.95}61.7 & \cellcolor[gray]{0.95}57.9 & \cellcolor[gray]{0.95}69.7 & \cellcolor[gray]{0.95}59.1 & 67.9 \\
  ORE\texttt{-(CC+EBUI)} \cite{joseph2021towards} & 65.1 & 74.6 & 57.9 & 39.5 & 36.7 & 75.1 & 80.1 & 73.3 & 37.1 & 69.8 & 48.6 & 69 & 77.5 & 72.8 & 76.5 & \cellcolor[gray]{0.95}34.4 & \cellcolor[gray]{0.95}62.6 & \cellcolor[gray]{0.95}56.5 & \cellcolor[gray]{0.95}80.3 & \cellcolor[gray]{0.95}65.7 & 62.6 \\
  ORE\texttt{-EBUI}\cite{joseph2021towards} & 75.4 & 81.1 & 62.7 & 51.3 & 55.7 & 77.2 & 85.6 & 81.7 & 46.1 & 76.2 & 55.4 & 76.7 & 86.2 & 78.5 & 82.1 & \cellcolor[gray]{0.95}32.8 & \cellcolor[gray]{0.95}63.6 & \cellcolor[gray]{0.95}54.7 & \cellcolor[gray]{0.95}77.7 & \cellcolor[gray]{0.95}64.6 & 68.5 \\
  OW-DETR \cite{gupta2022ow} & 77.1 & 76.5 & 69.2 & 51.3 & 61.3 & 79.8 & 84.2 & 81.0 & 49.7 & 79.6 & 58.1 & 79.9 & 83.1 & 67.8 & 85.4 & \cellcolor[gray]{0.95}33.2 & \cellcolor[gray]{0.95}65.1 & \cellcolor[gray]{0.95}62.0 & \cellcolor[gray]{0.95}73.9 & \cellcolor[gray]{0.95}65.0 & 69.4 \\
  PROB \cite{zohar2023prob} & 77.9 & 77.0 & 77.5 & 56.7 & 63.9 & 75.0 & 85.5 & 82.3 & 50.0 & 78.5 & 63.1 & 75.8 & 80.0 & 78.3 & 77.2 & \cellcolor[gray]{0.95}38.4 & \cellcolor[gray]{0.95}69.8 & \cellcolor[gray]{0.95}57.1 & \cellcolor[gray]{0.95}73.7 & \cellcolor[gray]{0.95}64.9 & \underline{70.1} \\ 
  \midrule
  \textbf{Ours: Decoupled PROB} & 79.6 & 77.0 & 74.9 & 57.8 & 62.8 & 77.0 & 84.9 & 84.4 & 46.4 & 73.8 & 64.9 & 76.4 & 80.6 & 77.6 & 76.7 & \cellcolor[gray]{0.95}41.8 & \cellcolor[gray]{0.95}72.6 & \cellcolor[gray]{0.95}61.1 & \cellcolor[gray]{0.95}74.5 & \cellcolor[gray]{0.95}65.4 & \textbf{70.5} \\ 
  \bottomrule
  \end{tabular}
  \end{adjustbox}
  
  \vspace{0.03cm}
  
  \begin{adjustbox}{width=0.97\linewidth,center}
  \begin{tabular}{lccccccccccccccccccccc}
  \toprule
  \textbf{\textcolor{green}{19 + 1 setting}} & aero & cycle & bird & boat & bottle & bus & car & cat & chair & cow & table & dog & horse & bike & person & plant & sheep & sofa & train & \cellcolor[gray]{0.95}tv & mAP \\
  \midrule
  ILOD \cite{shmelkov2017incremental} & 69.4 & 79.3 & 69.5 & 57.4 & 45.4 & 78.4 & 79.1 & 80.5 & 45.7 & 76.3 & 64.8 & 77.8 & 70.1 & 77.5 & 70.1 & 42.3 & 67.5 & 64.4 & 76.7 & \cellcolor[gray]{0.95}62.7 & 68.2 \\
  Faster ILOD \cite{peng2020faster} & 64.2 & 74.7 & 73.2 & 55.5 & 53.7 & 70.8 & 82.9 & 82.6 & 51.6 & 79.7 & 58.7 & 78.8 & 81.8 &75.3 & 77.4 & 43.1 & 73.8 & 61.7 & 69.8 & \cellcolor[gray]{0.95}61.1 & 68.5 \\
  ORE\texttt{-(CC+EBUI)} \cite{joseph2021towards} & 60.7 & 78.6 & 61.8 & 45.0 & 43.2 & 75.1 & 82.5 & 75.5 & 42.4 & 75.1 & 56.7 & 72.9 & 80.8 & 75.4 & 77.7 & 37.8 & 72.3 & 64.5 & 70.7 & \cellcolor[gray]{0.95}49.9 & 64.9 \\
  ORE\texttt{-EBUI}\cite{joseph2021towards} & 67.3 & 76.8 & 60.0 & 48.4 & 58.8 & 81.1 & 86.5 & 75.8 & 41.5 & 79.6 & 54.6 & 72.8 & 85.9 & 81.7 & 82.4 & 44.8 & 75.8 & 68.2 & 75.7 & \cellcolor[gray]{0.95}60.1 & 68.8 \\
  OW-DETR \cite{gupta2022ow} & 70.5 & 77.2 & 73.8 & 54.0 & 55.6 & 79.0 & 80.8 & 80.6 & 43.2 & 80.4 & 53.5 & 77.5 & 89.5 & 82.0 & 74.7 & 43.3 & 71.9 & 66.6 & 79.4 & \cellcolor[gray]{0.95}62.0 & 70.2 \\
  PROB \cite{zohar2023prob} & 80.3 & 78.9 & 77.6 & 59.7 & 63.7 & 75.2 & 86.0 & 83.9 & 53.7 & 82.8 & 66.5 & 82.7 & 80.6 & 83.8 &77.9 & 48.9 & 74.5 & 69.9 & 77.6 & \cellcolor[gray]{0.95}48.5 & \underline{72.6} \\  
  \midrule
  \textbf{Ours: Decoupled PROB} & 81.6 & 78.4 & 76.5 & 58.4 & 64.9 & 77.0 & 85.3 & 83.2 & 49.6 & 80.9 & 67.2 & 82.0 & 83.5 & 76.8 & 79.9 & 46.3 & 74.7 & 70.4 & 84.8 & \cellcolor[gray]{0.95}62.8  & \textbf{73.2} \\
  \bottomrule
  \end{tabular}
  \end{adjustbox}
  \label{tab:incremental}
\end{table*}

\subsection{State-of-the-Art Comparison}
\cref{tab:sota_comp} compares the model performance on M-OWODB and S-OWODB. 
Following \cite{gupta2022ow,zohar2023prob,ma2023cat}, for a fair comparison in the OWOD setting, we do not adopt the energy-based unknown identifier (EBUI) in ORE, 
which relies on weak unknown object supervision from holdout validation data.
Additionally, regarding USD, although it introduces an Auxiliary Supervision Framework (ASF) that utilizes the SAM\cite{kirillov2023segment}, we compare it to models without ASF for fairness. 
This is because ASF generates pseudo-GTs based on the predictions of a powerful external model, which other models do not utilize.

\noindent \textbf{M-OWODB.}
In the M-OWODB benchmark, we compare our method with the baseline model PROB and other top-performing models across multiple metrics. 
Decoupled PROB outperforms PROB in all metrics. 
Notably, the detection performance for known objects after Task 2 shows significant improvement, with ``Current known mAP'' surpassing PROB by 4.5\%, 3.9\%, and 3.2\% in Tasks 2, 3, and 4, respectively. 
Furthermore, when compared to other state-of-the-art methods such as CAT, OW-DETR, and USD, our model demonstrates superior performance in ``Current known mA'' and ``Both mAP'' from Task 2 onwards. 
It also achieves state-of-the-art performance in ``Previously known mAP'' from Task 3 onwards.
Further discussion on the results is provided in the supplementary material.

\noindent \textbf{S-OWODB.}
In the S-OWODB benchmark, Decoupled PROB outperforms PROB in all metrics except for ``U-Recall'' in Task 1, ``Previous Known mAP'' in Task2, and ``U-Recall'' in Task 3. 
Similar to M-OWODB, Decoupled PROB significantly surpasses PROB in ``Current known mAP'' from Task 2 onwards, with improvements of 9.7\%, 12.9\%, and 8.1\% in Tasks 2, 3, and 4, respectively. 
Additionally, compared to other state-of-the-art methods, our model excels in many metrics, showing state-of-the-art performance ``U-Recall'' in Task 2, ``Current known mAP'' and ``Both mAP'' in Task 2 onwards, and ``Previously known mAP'' in Task 4 .
Further discussion on the results is provided in the supplementary material.

\noindent \textbf{Qualitative Results.}
\cref{fig:comp} shows a qualitative comparison on test set images. 
The right two images demonstrate that Decoupled PROB (top row) correctly detects objects as known classes (green) that PROB (bottom row) detects as unknown classes (yellow). 
Specifically, Decoupled PROB accurately detects the umbrella and baseball glove as known classes in the left-most and second-to-left images, respectively. 
Additionally, the right two images show that Decoupled PROB is able to detect unknown objects that PROB fails to detect.

\subsection{Incremental Object Detection}
As experimented in \cite{gupta2022ow,zohar2023prob}, we compare our method with other approaches for the incremental object detection task. 
This task involves three standard settings on Pascal VOC 2007. The model first learns 10, 15, and 19 object classes, and then additional 10, 5, and 1 classes are incrementally introduced. 
As shown in \cref{tab:incremental}, Decoupled PROB demonstrates favorable performance compared to existing methods and the baseline model, PROB. 

\label{sec:experiment}

\section{Conclusion}
We focused on the task of Open World Object Detection, analyzing and addressing the issues of learning conflicts between objectness and class predictions in PROB, as well as the initialization of object queries. 
For query initialization, we introduced Task-Decoupled Query Initialization (TDQI), which combines query selection and learnable queries, enabling effective and efficient feature extraction around objects in the decoder layers. 
TDQI allows the decoder to capture object features in the shallow layers, facilitating the implementation of Early Termination of Objectness Prediction (ETOP). 
ETOP stops objectness estimation in the shallow layers of the decoder, mitigating the conflict between class and objectness learning. 
Notably, TDQI can be easily integrated into other Deformable DETR-based OWOD models. 
By incorporating these features, our proposed model, Decoupled PROB, demonstrated state-of-the-art results on several metrics in OWOD benchmarks.

\label{sec:conclusion}

\maketitlesupplementary
\section{Introduction}

We report the training loss (\cref{sec:tr_loss}), additional details of the training (\cref{sec:add_tr}), ablation study (\cref{sec:ab_st}), comparison between ETOP and DOL (\cref{sec:etop_dol}), 
further discussion on the results (\cref{sec:further}), analysis of TDQI (\cref{sec:an_tq}), comparison of model size and computational cost (\cref{sec:cost}), additional qualitative results (\cref{sec:add_qr}), and limitations (\cref{sec:limi}) in the supplementary material.

\section{Training Loss} \label{sec:tr_loss}
Decoupled PROB is trained with the following loss function:
\begin{align}
L = L_{bb} + L_{cls} + L_{obj}
\end{align}
where $L_{bb}$ denotes the $L_1$ and gIoU losses for bounding box learning, $L_{cls}$ represents the sigmoid focal loss for class classification, and $L_{obj}$ is the objectness loss(explained in main paper). 
All these settings are identical to those used in PROB\cite{zohar2023prob}.

\section{Additional Training Details} \label{sec:add_tr}
We report additional training details for Decoupled PROB that are not included in the main paper. 
\cref{tab:training_hyperparameters} lists the number of training epochs and the epochs at which the learning rate drops for each task in M-OWODB and S-OWODB. 
`` - '' in the learning rate drop column indicates that no special settings are applied. All other hyperparameters are the same as those used in PROB\cite{zohar2023prob}.
We use two Nvidia RTX A6000 GPUs for training, setting the batch size to 6 for each GPU. 

\begin{table*}[tp]
    \centering
    \caption{\textbf{Details for training hyperparameters.}}
    \begin{adjustbox}{width=0.97\linewidth,center}
    \begin{tabular}{l|cc|cc|cc|cc|cc|cc|cc}
    \toprule
    Training Session & \multicolumn{2}{c|}{\textbf{Task 1}} & \multicolumn{2}{c|}{\textbf{Task 2}} & \multicolumn{2}{c|}{\textbf{Task 2 ft}} & 
    \multicolumn{2}{c|}{\textbf{Task 3}} & \multicolumn{2}{c|}{\textbf{Task 3 ft}} & \multicolumn{2}{c|}{\textbf{Task 4}} & \multicolumn{2}{c}{\textbf{Task 4 ft}} \\
    & Epochs & lr drop & Epochs & lr drop & Epochs & lr drop & Epochs & lr drop & Epochs & lr drop & Epochs & lr drop & Epochs & lr drop  \\
    \midrule
    M-OWODB & 26 & 15 & 10 & - & 15 & 10 & 10 & - & 15 & 10 & 10 & - & 15 & 10 \\
    S-OWODB & 26 & 15 & 10 & - & 10 & 5 & 10 & - & 10 & 5 & 10 & - & 10 & 5 \\
    \bottomrule
    \end{tabular}
    \end{adjustbox}
    \label{tab:training_hyperparameters}
\end{table*}

\section{Ablation Study} \label{sec:ab_st}
\cref{tab:sota_comp} shows the ablation study for Decoupled PROB. 
In \textbf{Decoupled PROB}\texttt{-TDQI}$^{*1}$, we replace Task-Decoupled Query Initialization (TDQI) in Decoupled PROB with a simple class score-based query selection\cite{zhu2020deformable,yao2021efficient}. 
In \textbf{Decoupled PROB}\texttt{-TDQI}$^{*2}$, all object queries use only learnable parameters instead of TDQI. 
In \textbf{Decoupled PROB}\texttt{-ETOP}, we remove Early Termination of Objectness Prediction (ETOP) from Decoupled PROB, 
performing objectness prediction until the last layer of the decoder and learning without mitigating the conflict between objectness and class predictions. 
As shown in \cref{tab:sota_comp}, \textbf{Decoupled PROB}\texttt{-TDQI}$^{*1}$ exhibits lower performance across all metrics compared to Decoupled PROB, highlighting the importance of TDQI in Decoupled PROB. 
However, it should be noted that, similar to our proposed method, the objectness prediction is stopped at decoder layer 2(ETOP), which is the optimal setting for TDQI. 
Exploring the optimal number of layers for class score-based query selection could potentially improve performance.
While \textbf{Decoupled PROB}\texttt{-TDQI}$^{*2}$ improves performance for unknown objects, its performance on known objects significantly decreases compared to Decoupled PROB. 
As shown in Table 2 of the main paper, it can be observed that the higher the number of learnable parameters, the better the performance on unknown objects, and \textbf{Decoupled PROB}\texttt{-TDQI}$^{*2}$ follows this trend.
Similar to \textbf{Decoupled PROB}\texttt{-TDQI}$^{*1}$, it should be noted that \textbf{Decoupled PROB}\texttt{-TDQI}$^{*2}$ could potentially improve performance by exploring the optimal number of layers to stop objectness prediction in ETOP.
Regarding \textbf{Decoupled PROB}\texttt{-ETOP}, while it maintains comparable known object detection performance to Decoupled PROB, it shows significantly lower performance in unknown object detection. 
This underscores the crucial role of ETOP in enhancing the performance of Decoupled PROB.

\begin{table*}[t]
    \centering
    \caption{\textbf{Comparison of ablation experiment results for the components of the proposed model.}
    The details of the metrics are provided in Section 5 of the main paper. 
    In \textbf{Decoupled PROB}\texttt{-TDQI}$^{*1}$, we modify TDQI to use a purely class score-based query selection. 
    In \textbf{Decoupled PROB}\texttt{-TDQI}$^{*2}$, we modify TDQI to use only learnable parameters.
    In \textbf{Decoupled PROB}\texttt{-ETOP}, objectness is predicted and trained until the last layer of the decoder.
    The performance comparison also includes Deformable DETR and the Upper Bound, which is Deformable DETR trained with ground truth for unknown classes, as reported in \cite{gupta2022ow}.}
    \begin{adjustbox}{width=0.97\linewidth,center}
      \begin{tabular}{
        l|
        >{\columncolor{yellow!20}}c
        c
        c|
        >{\columncolor{yellow!20}}c
        c
        c
        c|
        >{\columncolor{yellow!20}}c
        c
        c
        c|
        c
        c
        c}
        \toprule
        \textbf{Task IDs (→)} & \multicolumn{3}{c|}{\textbf{Task 1}} & \multicolumn{4}{c|}{\textbf{Task 2}} & \multicolumn{4}{c|}{\textbf{Task 3}} & \multicolumn{3}{c}{\textbf{Task 4}} \\
        \midrule
        & U-Recall & \multicolumn{2}{c|}{\cellcolor{cyan!10}mAP (↑)} & U-Recall & \multicolumn{3}{c|}{\cellcolor{cyan!10}mAP (↑)} & U-Recall & \multicolumn{3}{c|}{\cellcolor{cyan!10}mAP (↑)} & \multicolumn{3}{c}{\cellcolor{cyan!10}mAP (↑)} \\
        & (↑) & \multicolumn{2}{c|}{\shortstack{Current \\ known}} & (↑) & \shortstack{Previously \\ known} & \shortstack{Current \\ known} & Both & (↑) & \shortstack{Previously \\ known} & \shortstack{Current \\ known} & Both & \shortstack{Previously \\ known} & \shortstack{Current \\ known} & Both \\
        \midrule
        Upper Bound & 31.6 & \multicolumn{2}{c|}{62.5} & 40.5 & 55.8 & 38.1 & 46.9 & 42.6 & 42.4 & 29.3 & 33.9 & 35.6 & 23.1 & 32.5 \\
        D-DETR \cite{zhu2020deformable} & - & \multicolumn{2}{c|}{60.3} & - & 54.5 & 34.4 & 44.7 & - & 40.0 & 17.7 & 33.3 & 32.5 & 20.0 & 29.4 \\
        \midrule
        \midrule
        \textbf{Decoupled PROB}\texttt{-TDQI}$^{*1}$ & 17.9 & \multicolumn{2}{c|}{57.4} & 17.5 & 55.1 & 36.0 & 45.6 & \underline{18.8} & 44.1 & 25.5 & 37.9 & 36.9 & 21.2 & 33.0 \\
        \textbf{Decoupled PROB}\texttt{-TDQI}$^{*2}$ & \textbf{20.9} & \multicolumn{2}{c|}{59.3} & \textbf{18.6} & 55.6 & 36.6 & 46.1 & \textbf{21.0} & 44.2 & 24.6 & 37.7 & 37.2 & 21.7 & 33.3 \\
        \textbf{Decoupled PROB}\texttt{-ETOP} & 18.8 & \multicolumn{2}{c|}{\textbf{60.4}} & 15.3 & \textbf{56.4} & \textbf{37.2} & \textbf{46.8} & 17.9 & \textbf{44.7} & \underline{25.7} & \underline{38.4} & \underline{37.2} & \textbf{22.8} & \underline{33.6} \\
        Final: \textbf{Decoupled PROB} & \underline{20.3} & \multicolumn{2}{c|}{\underline{59.8}} & \underline{18.4} & \textbf{56.4} & \underline{36.7} & \underline{46.6} & \underline{20.3} & \underline{44.6} & \textbf{26.1} & \textbf{38.5} & \textbf{37.8} & \underline{22.1} & \textbf{33.8} \\
        \bottomrule
      \end{tabular}
    \end{adjustbox}
  \label{tab:sota_comp}
  \end{table*}

\section{Comparison between ETOP and DOL} \label{sec:etop_dol}
Unlike Decoupled Objectness Learning (DOL)\cite{he2023usd}, ETOP performs both class and bounding box prediction concurrently in the decoder layers that predict objectness. 
Additionally, ETOP incorporates iterative refinement for bounding box prediction. The combination of ETOP and TDQI offers the advantage of enabling iterative refinement across a greater number of layers through query selection.

\cref{tab:etop_dol} presents the experimental results comparing ETOP and DOL. 
TDQI+DOL indicates that ETOP in Decoupled PROB is replaced with DOL.
Additionally, DOL$^{*1}$ stops objectness prediction at the first decoder layer, while DOL$^{*2}$ stops it at the second decoder layer. 
As shown in the \cref{tab:etop_dol}, using ETOP achieves the best performance across all metrics. 
In TDQI, bounding box prediction is also performed during query selection, which allows iterative refinement over more layers based on those coordinates. 
This likely explains why ETOP demonstrates superior performance compared to DOL.

Furthermore, these results highlight the advantages of ETOP's continuous class and bounding box prediction, suggesting that it is not necessary to have layers dedicated solely to objectness prediction in Decoupled PROB.

  \begin{table*}[t]
    \centering
    \caption{\textbf{Comparison between ETOP and DOL.}
    DOL$^{*1}$ and DOL$^{*2}$ indicate that the objectness prediction in DOL stops at the first decoder layer and the second decoder layer, respectively.}
    \begin{adjustbox}{width=0.97\linewidth,center}
      \begin{tabular}{
        l|
        >{\columncolor{yellow!20}}c
        c
        c|
        >{\columncolor{yellow!20}}c
        c
        c
        c|
        >{\columncolor{yellow!20}}c
        c
        c
        c|
        c
        c
        c}
        \toprule
        \textbf{Task IDs (→)} & \multicolumn{3}{c|}{\textbf{Task 1}} & \multicolumn{4}{c|}{\textbf{Task 2}} & \multicolumn{4}{c|}{\textbf{Task 3}} & \multicolumn{3}{c}{\textbf{Task 4}} \\
        \midrule
        & U-Recall & \multicolumn{2}{c|}{\cellcolor{cyan!10}mAP (↑)} & U-Recall & \multicolumn{3}{c|}{\cellcolor{cyan!10}mAP (↑)} & U-Recall & \multicolumn{3}{c|}{\cellcolor{cyan!10}mAP (↑)} & \multicolumn{3}{c}{\cellcolor{cyan!10}mAP (↑)} \\
        & (↑) & \multicolumn{2}{c|}{\shortstack{Current \\ known}} & (↑) & \shortstack{Previously \\ known} & \shortstack{Current \\ known} & Both & (↑) & \shortstack{Previously \\ known} & \shortstack{Current \\ known} & Both & \shortstack{Previously \\ known} & \shortstack{Current \\ known} & Both \\
        \midrule
        TDQI + DOL$^{*1}$ & 19.2 & \multicolumn{2}{c|}{59.1} & 17.4 & 53.4 & 33.9 & 43.6 & 19.2 & 43.0 & 23.3 & 36.4 & 36.5 & 20.7 & 32.5 \\
        TDQI + DOL$^{*2}$ & 19.8 & \multicolumn{2}{c|}{58.2} & 17.0 & 53.4 & 34.0 & 43.7 & 20.0 & 42.2 & 23.7 & 36.0 & 36.0 & 20.6 & 32.1 \\
        \textbf{Decoupled PROB} & \textbf{20.3} & \multicolumn{2}{c|}{\textbf{59.8}} & \textbf{18.4} & \textbf{56.4} & \textbf{36.7} & \textbf{46.6} & \textbf{20.3} & \textbf{44.6} & \textbf{26.1} & \textbf{38.5} & \textbf{37.8} & \textbf{22.1} & \textbf{33.8} \\
        \bottomrule
      \end{tabular}
    \end{adjustbox}
  \label{tab:etop_dol}
  \end{table*}

\section{Further Discussion on the Results} \label{sec:further}
As shown in Table 4 of the main paper, our model often underperforms compared to CAT and USD\texttt{-ASF}, particularly in terms of U-Recall, on both the M-OWOD and S-OWOD benchmarks. 
This may be influenced by the number of object queries initialized as learnable queries. 
As indicated in Table 2 of the main paper, a higher ratio of learnable queries tends to improve U-Recall. 
Both CAT and USD\texttt{-ASF} initialize all object queries as learnable queries. 
In our approach, we use 20 object queries initialized by query selection and 80 by learnable queries.

Another reason CAT and USD\texttt{-ASF} may show superior metrics in early-stage tasks could be that the learnable queries in our proposed model are not fully converged yet. 
While our method utilizes both query selection and learnable queries, query selection is likely to focus on object surroundings from the early stages of the decoder, leading to faster convergence in training compared to learnable queries. 
This suggests that hyperparameter settings for our method might be more complex compared to those using only learnable queries. 
Although our proposed model follows the settings outlined in \cref{tab:training_hyperparameters} of the supplementary material, 
USD\texttt{-ASF} and CAT undergo longer training periods (e.g., USD trains for 41 epochs and CAT for 45 epochs in task 1).
Particularly in early-stage tasks such as task 1 and task 2, there is a possibility that the learnable query-based object queries in our model are not fully converged.

Additionally, one of the reasons for CAT's superior U-Recall could be that the pseudo-labeling method it uses, based on selective search, functions exceptionally well.

\section{Analysis of Task-Decoupled Query Initialization (TDQI)} \label{sec:an_tq}
\begin{figure}[tp]
  \centering
  \includegraphics[width=0.96\linewidth]{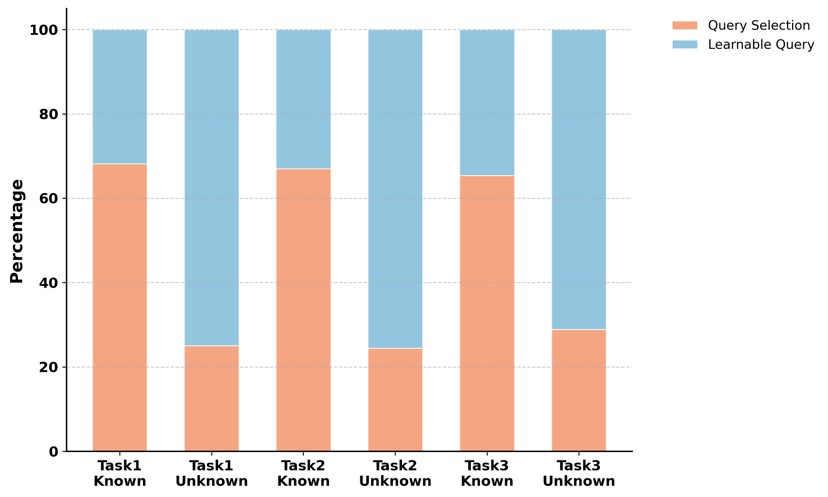}
  \caption{\textbf{Comparison of known class detection and unknown class detection roles in each task for TDQI.
  }}
  \label{fig:task_lq_qs}
\end{figure}

In TDQI, the object queries initialized with query selection are responsible for detecting known objects, 
while those initialized with learnable query cover missed known objects and unknown objects. 
We investigated the detection ratios of known and unknown classes in each task of OWODB using TDQI. 
The results for M-OWODB are shown in \cref{fig:task_lq_qs}. 
As in the main paper, 20 object queries are initialized with query selection, and 80 object queries are initialized with learnable query. 

Although object queries initialized with query selection account for only about 20\% of the total, they are responsible for nearly 70\% of the detection of known classes. 
Conversely, learnable queries are responsible for nearly 70\% of the detection of unknown objects. 
This demonstrates that in TDQI, object queries initialized with query selection primarily handle the detection of known objects, 
while those initialized with learnable query cover the missed known objects and unknown objects.

\section{Comparison of Model Size and Computational Cost.} \label{sec:cost}
\cref{tab:size_and_cost} presents the number of parameters and the computational cost for each OWOD model. 
The computational cost is calculated with an image size of $640\times640$.
OW-DETR\cite{gupta2022ow}, CAT\cite{ma2023cat}, PROB\cite{zohar2023prob}, and Decoupled PROB, which are based on the Deformable DETR model\cite{zhu2020deformable}, have nearly the same number of parameters and computational costs.

OrthogonalDet\cite{sun2024exploring}, on the other hand, is a recently proposed high-performance OWOD model. 
This model has a significantly larger number of parameters and higher computational cost, and in our implementation, it outputs nearly 2000 detections (refer to the supplementary material of \cite{sun2024exploring}) compared to the 100 detections output by models such as PROB and Decoupled PROB. 
For more details on the performance of OrthogonalDet, please refer to \cite{sun2024exploring}.

\begin{table}[tp]
\centering
\caption{\textbf{Comparison of model size and computational cost.}}
\begin{adjustbox}{width=0.97\linewidth,center}
\begin{tabular}{c|cc}
  \toprule
  \textbf{Methods} & \textbf{Params} & \textbf{FLOPs} \\
  \midrule
  OWOD \cite{joseph2021towards} & 33.6M & 32.1G \\
  OW-DETR \cite{gupta2022ow} & 39.7M & 156.3G \\
  CAT \cite{ma2023cat} & 46.1M & 164.3G \\
  PROB \cite{zohar2023prob} & 39.7M & 156.3G \\
  OrthogonalDet \cite{sun2024exploring} \* & 106.0M & 1616.7G \\
  Decoupled PROB (Ours) & 40.9M & 163.4G \\
  \bottomrule
\end{tabular}
\end{adjustbox}
\label{tab:size_and_cost}
\end{table}

\section{Additional Qualitative Results} \label{sec:add_qr}
\begin{figure*}[tp]
  \centering
  \includegraphics[height=0.35\textwidth, width=1.\linewidth]{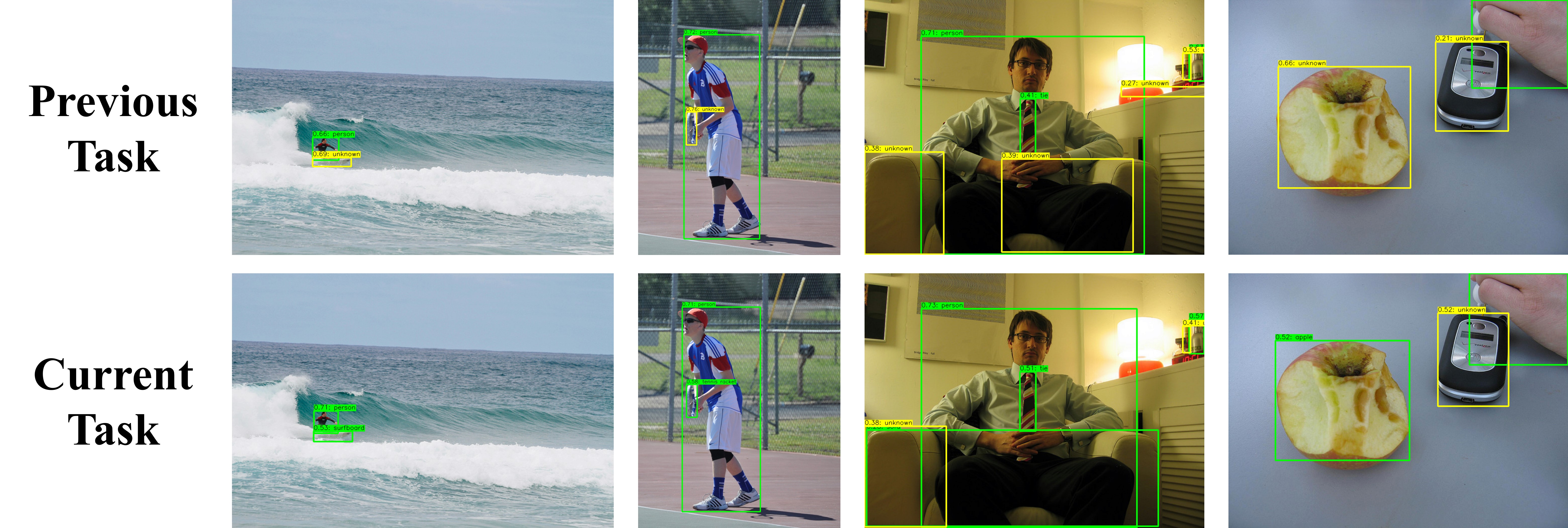}
  \caption{\textbf{Learning Unknown Classes as Known Classes (Incremental Learning).}}
  \label{fig:incre_imgs}
\end{figure*}

\cref{fig:incre_imgs} illustrates examples where objects that were labeled as unknown classes in previous tasks are provided to Decoupled PROB in the current task, allowing it to learn and detect them as known classes. 
From left to right, the surfboard, tennis racket, sofa, and apple are detected as known classes, having transitioned from unknown classes.

\begin{figure*}[tp]
  \centering
  \includegraphics[width=0.96\linewidth]{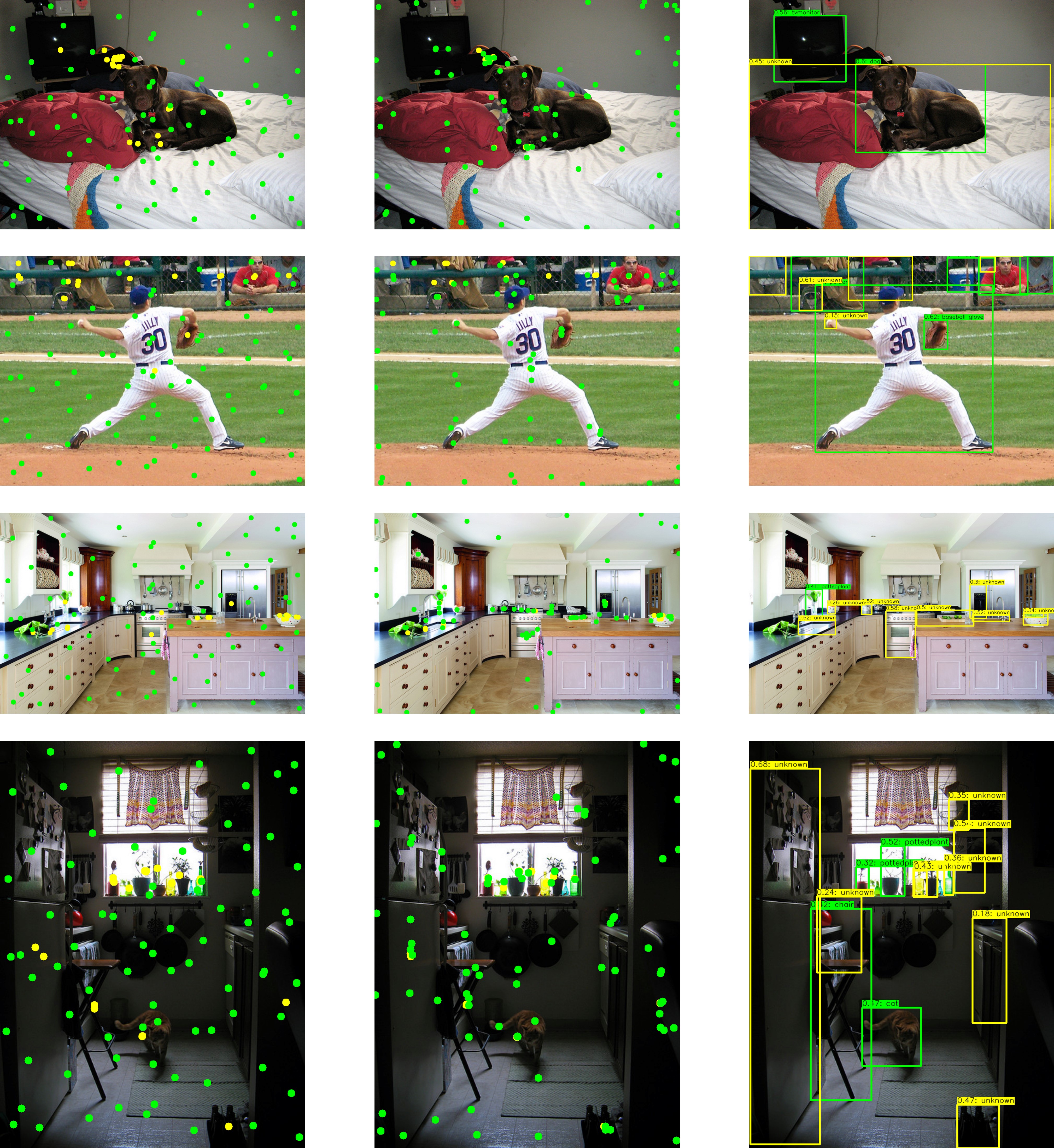}
  \caption{\textbf{Qualitative results on example images from test set.}}
  \label{fig:pred_imgs}
\end{figure*}

\cref{fig:pred_imgs} illustrates the qualitative results of Decoupled PROB. 
From left to right, it shows the reference points in the initial layer of the decoder, the reference points in the last layer of the decoder, and the detection results. 
The yellow reference points correspond to object queries initialized by query selection, while the green reference points correspond to object queries initialized by learnable query. 
The green bounding boxes indicate known classes, and the yellow bounding boxes indicate unknown classes.

\section{Limitations} \label{sec:limi}
As shown in the results of the main paper, OWOD research has been rapidly advancing. 
While this task shows promise for applications in fields such as autonomous driving and robotics, there are still performance improvements needed, including in our model. 
These improvements include reducing false detections of background as unknown objects, improving the accurate localization of unknown objects, and preventing the forgetting of known classes.

PROB\cite{zohar2023prob}, which we used as our baseline, is a remarkable approach. 
However, it may have an issue where background information is included in the query embeddings used to update the objectness distribution, indicating a need for future improvements. 
Developing new representation methods to distinguish between objects and background remains an important research challenge. Further studies are required to enable models to fundamentally understand what constitutes an object.
\label{sec:Appendix}

{\small
\bibliographystyle{ieee_fullname}
\bibliography{egbib}

\begin{thebibliography}{10}\itemsep=-1pt

\bibitem{carion2020end}
Nicolas Carion, Francisco Massa, Gabriel Synnaeve, Nicolas Usunier, Alexander Kirillov, and Sergey Zagoruyko.
\newblock End-to-end object detection with transformers.
\newblock In {\em European conference on computer vision}, pages 213--229. Springer, 2020.

\bibitem{caron2021emerging}
Mathilde Caron, Hugo Touvron, Ishan Misra, Herv{\'e} J{\'e}gou, Julien Mairal, Piotr Bojanowski, and Armand Joulin.
\newblock Emerging properties in self-supervised vision transformers.
\newblock In {\em Proceedings of the IEEE/CVF international conference on computer vision}, pages 9650--9660, 2021.

\bibitem{chen2020learning}
Guangyao Chen, Limeng Qiao, Yemin Shi, Peixi Peng, Jia Li, Tiejun Huang, Shiliang Pu, and Yonghong Tian.
\newblock Learning open set network with discriminative reciprocal points.
\newblock In {\em Computer Vision--ECCV 2020: 16th European Conference, Glasgow, UK, August 23--28, 2020, Proceedings, Part III 16}, pages 507--522. Springer, 2020.

\bibitem{chen2023group}
Qiang Chen, Xiaokang Chen, Jian Wang, Shan Zhang, Kun Yao, Haocheng Feng, Junyu Han, Errui Ding, Gang Zeng, and Jingdong Wang.
\newblock Group detr: Fast detr training with group-wise one-to-many assignment.
\newblock In {\em Proceedings of the IEEE/CVF International Conference on Computer Vision}, pages 6633--6642, 2023.

\bibitem{dai2021dynamic}
Xiyang Dai, Yinpeng Chen, Jianwei Yang, Pengchuan Zhang, Lu Yuan, and Lei Zhang.
\newblock Dynamic detr: End-to-end object detection with dynamic attention.
\newblock In {\em Proceedings of the IEEE/CVF international conference on computer vision}, pages 2988--2997, 2021.

\bibitem{dosovitskiy2020image}
Alexey Dosovitskiy, Lucas Beyer, Alexander Kolesnikov, Dirk Weissenborn, Xiaohua Zhai, Thomas Unterthiner, Mostafa Dehghani, Matthias Minderer, Georg Heigold, Sylvain Gelly, et~al.
\newblock An image is worth 16x16 words: Transformers for image recognition at scale.
\newblock In {\em International Conference on Learning Representations}, 2020.

\bibitem{du2022unknown}
Xuefeng Du, Xin Wang, Gabriel Gozum, and Yixuan Li.
\newblock Unknown-aware object detection: Learning what you don't know from videos in the wild.
\newblock In {\em Proceedings of the IEEE/CVF Conference on Computer Vision and Pattern Recognition}, pages 13678--13688, 2022.

\bibitem{everingham2010pascal}
Mark Everingham, Luc Van~Gool, Christopher~KI Williams, John Winn, and Andrew Zisserman.
\newblock The pascal visual object classes (voc) challenge.
\newblock {\em International journal of computer vision}, 88:303--338, 2010.

\bibitem{gupta2022ow}
Akshita Gupta, Sanath Narayan, KJ Joseph, Salman Khan, Fahad~Shahbaz Khan, and Mubarak Shah.
\newblock Ow-detr: Open-world detection transformer.
\newblock In {\em Proceedings of the IEEE/CVF conference on computer vision and pattern recognition}, pages 9235--9244, 2022.

\bibitem{he2016deep}
Kaiming He, Xiangyu Zhang, Shaoqing Ren, and Jian Sun.
\newblock Deep residual learning for image recognition.
\newblock In {\em Proceedings of the IEEE conference on computer vision and pattern recognition}, pages 770--778, 2016.

\bibitem{he2023usd}
Yulin He, Wei Chen, Yusong Tan, and Siqi Wang.
\newblock Usd: Unknown sensitive detector empowered by decoupled objectness and segment anything model.
\newblock {\em arXiv preprint arXiv:2306.02275}, 2023.

\bibitem{joseph2021towards}
KJ Joseph, Salman Khan, Fahad~Shahbaz Khan, and Vineeth~N Balasubramanian.
\newblock Towards open world object detection.
\newblock In {\em Proceedings of the IEEE/CVF conference on computer vision and pattern recognition}, pages 5830--5840, 2021.

\bibitem{kirillov2023segment}
Alexander Kirillov, Eric Mintun, Nikhila Ravi, Hanzi Mao, Chloe Rolland, Laura Gustafson, Tete Xiao, Spencer Whitehead, Alexander~C Berg, Wan-Yen Lo, et~al.
\newblock Segment anything.
\newblock In {\em Proceedings of the IEEE/CVF International Conference on Computer Vision}, pages 4015--4026, 2023.

\bibitem{kumar2023normalizing}
Nishant Kumar, Sini{\v{s}}a {\v{S}}egvi{\'c}, Abouzar Eslami, and Stefan Gumhold.
\newblock Normalizing flow based feature synthesis for outlier-aware object detection.
\newblock In {\em Proceedings of the IEEE/CVF Conference on Computer Vision and Pattern Recognition}, pages 5156--5165, 2023.

\bibitem{li2023lite}
Feng Li, Ailing Zeng, Shilong Liu, Hao Zhang, Hongyang Li, Lei Zhang, and Lionel~M Ni.
\newblock Lite detr: An interleaved multi-scale encoder for efficient detr.
\newblock In {\em Proceedings of the IEEE/CVF Conference on Computer Vision and Pattern Recognition}, pages 18558--18567, 2023.

\bibitem{li2022dn}
Feng Li, Hao Zhang, Shilong Liu, Jian Guo, Lionel~M Ni, and Lei Zhang.
\newblock Dn-detr: Accelerate detr training by introducing query denoising.
\newblock In {\em Proceedings of the IEEE/CVF conference on computer vision and pattern recognition}, pages 13619--13627, 2022.

\bibitem{liang2023unknown}
Wenteng Liang, Feng Xue, Yihao Liu, Guofeng Zhong, and Anlong Ming.
\newblock Unknown sniffer for object detection: Don't turn a blind eye to unknown objects.
\newblock In {\em Proceedings of the IEEE/CVF Conference on Computer Vision and Pattern Recognition}, pages 3230--3239, 2023.

\bibitem{lin2014microsoft}
Tsung-Yi Lin, Michael Maire, Serge Belongie, James Hays, Pietro Perona, Deva Ramanan, Piotr Doll{\'a}r, and C~Lawrence Zitnick.
\newblock Microsoft coco: Common objects in context.
\newblock In {\em Computer Vision--ECCV 2014: 13th European Conference, Zurich, Switzerland, September 6-12, 2014, Proceedings, Part V 13}, pages 740--755. Springer, 2014.

\bibitem{liu2021dab}
Shilong Liu, Feng Li, Hao Zhang, Xiao Yang, Xianbiao Qi, Hang Su, Jun Zhu, and Lei Zhang.
\newblock Dab-detr: Dynamic anchor boxes are better queries for detr.
\newblock In {\em International Conference on Learning Representations}, 2021.

\bibitem{loshchilov2018decoupled}
Ilya Loshchilov and Frank Hutter.
\newblock Decoupled weight decay regularization.
\newblock In {\em International Conference on Learning Representations}, 2018.

\bibitem{ma2023cat}
Shuailei Ma, Yuefeng Wang, Ying Wei, Jiaqi Fan, Thomas~H Li, Hongli Liu, and Fanbing Lv.
\newblock Cat: Localization and identification cascade detection transformer for open-world object detection.
\newblock In {\em Proceedings of the IEEE/CVF Conference on Computer Vision and Pattern Recognition}, pages 19681--19690, 2023.

\bibitem{ma2022rethinking}
Zeyu Ma, Yang Yang, Guoqing Wang, Xing Xu, Heng~Tao Shen, and Mingxing Zhang.
\newblock Rethinking open-world object detection in autonomous driving scenarios.
\newblock In {\em Proceedings of the 30th ACM International Conference on Multimedia}, pages 1279--1288, 2022.

\bibitem{meng2021conditional}
Depu Meng, Xiaokang Chen, Zejia Fan, Gang Zeng, Houqiang Li, Yuhui Yuan, Lei Sun, and Jingdong Wang.
\newblock Conditional detr for fast training convergence.
\newblock In {\em Proceedings of the IEEE/CVF international conference on computer vision}, pages 3651--3660, 2021.

\bibitem{peng2020faster}
Can Peng, Kun Zhao, and Brian~C Lovell.
\newblock Faster ilod: Incremental learning for object detectors based on faster rcnn.
\newblock {\em Pattern recognition letters}, 140:109--115, 2020.

\bibitem{pu2024rank}
Yifan Pu, Weicong Liang, Yiduo Hao, Yuhui Yuan, Yukang Yang, Chao Zhang, Han Hu, and Gao Huang.
\newblock Rank-detr for high quality object detection.
\newblock {\em Advances in Neural Information Processing Systems}, 36, 2024.

\bibitem{ren2016faster}
Shaoqing Ren, Kaiming He, Ross Girshick, and Jian Sun.
\newblock Faster r-cnn: Towards real-time object detection with region proposal networks.
\newblock {\em IEEE transactions on pattern analysis and machine intelligence}, 39(6):1137--1149, 2016.

\bibitem{roh2021sparse}
Byungseok Roh, JaeWoong Shin, Wuhyun Shin, and Saehoon Kim.
\newblock Sparse detr: Efficient end-to-end object detection with learnable sparsity.
\newblock {\em arXiv preprint arXiv:2111.14330}, 2021.

\bibitem{saito2022learning}
Kuniaki Saito, Ping Hu, Trevor Darrell, and Kate Saenko.
\newblock Learning to detect every thing in an open world.
\newblock In {\em European Conference on Computer Vision}, pages 268--284. Springer, 2022.

\bibitem{shmelkov2017incremental}
Konstantin Shmelkov, Cordelia Schmid, and Karteek Alahari.
\newblock Incremental learning of object detectors without catastrophic forgetting.
\newblock In {\em Proceedings of the IEEE international conference on computer vision}, pages 3400--3409, 2017.

\bibitem{singh2021order}
Deepak~Kumar Singh, Shyam~Nandan Rai, KJ Joseph, Rohit Saluja, Vineeth~N Balasubramanian, Chetan Arora, Anbumani Subramanian, and CV Jawahar.
\newblock Order: Open world object detection on road scenes.
\newblock In {\em Proc. NeurIPS Workshops}, volume~1, page~3, 2021.

\bibitem{sun2024exploring}
Zhicheng Sun, Jinghan Li, and Yadong Mu.
\newblock Exploring orthogonality in open world object detection.
\newblock In {\em Proceedings of the IEEE/CVF Conference on Computer Vision and Pattern Recognition}, pages 17302--17312, 2024.

\bibitem{uijlings2013selective}
Jasper~RR Uijlings, Koen~EA Van De~Sande, Theo Gevers, and Arnold~WM Smeulders.
\newblock Selective search for object recognition.
\newblock {\em International journal of computer vision}, 104:154--171, 2013.

\bibitem{wang2022anchor}
Yingming Wang, Xiangyu Zhang, Tong Yang, and Jian Sun.
\newblock Anchor detr: Query design for transformer-based detector.
\newblock In {\em Proceedings of the AAAI conference on artificial intelligence}, volume~36, pages 2567--2575, 2022.

\bibitem{wu2023discriminating}
Aming Wu and Cheng Deng.
\newblock Discriminating known from unknown objects via structure-enhanced recurrent variational autoencoder.
\newblock In {\em Proceedings of the IEEE/CVF Conference on Computer Vision and Pattern Recognition}, pages 23956--23965, 2023.

\bibitem{wu2022two}
Yan Wu, Xiaowei Zhao, Yuqing Ma, Duorui Wang, and Xianglong Liu.
\newblock Two-branch objectness-centric open world detection.
\newblock In {\em Proceedings of the 3rd International Workshop on Human-Centric Multimedia Analysis}, pages 35--40, 2022.

\bibitem{wu2022uc}
Zhiheng Wu, Yue Lu, Xingyu Chen, Zhengxing Wu, Liwen Kang, and Junzhi Yu.
\newblock Uc-owod: Unknown-classified open world object detection.
\newblock In {\em European Conference on Computer Vision}, pages 193--210. Springer, 2022.

\bibitem{yang2020convolutional}
Hong-Ming Yang, Xu-Yao Zhang, Fei Yin, Qing Yang, and Cheng-Lin Liu.
\newblock Convolutional prototype network for open set recognition.
\newblock {\em IEEE Transactions on Pattern Analysis and Machine Intelligence}, 44(5):2358--2370, 2020.

\bibitem{yao2021efficient}
Zhuyu Yao, Jiangbo Ai, Boxun Li, and Chi Zhang.
\newblock Efficient detr: improving end-to-end object detector with dense prior.
\newblock {\em arXiv preprint arXiv:2104.01318}, 2021.

\bibitem{yu2022open}
Jinan Yu, Liyan Ma, Zhenglin Li, Yan Peng, and Shaorong Xie.
\newblock Open-world object detection via discriminative class prototype learning.
\newblock In {\em 2022 IEEE International Conference on Image Processing (ICIP)}, pages 626--630. IEEE, 2022.

\bibitem{zhang2022dino}
Hao Zhang, Feng Li, Shilong Liu, Lei Zhang, Hang Su, Jun Zhu, Lionel Ni, and Heung-Yeung Shum.
\newblock Dino: Detr with improved denoising anchor boxes for end-to-end object detection.
\newblock In {\em The Eleventh International Conference on Learning Representations}, 2022.

\bibitem{zhao2023revisiting}
Xiaowei Zhao, Yuqing Ma, Duorui Wang, Yifan Shen, Yixuan Qiao, and Xianglong Liu.
\newblock Revisiting open world object detection.
\newblock {\em IEEE Transactions on Circuits and Systems for Video Technology}, 2023.

\bibitem{zhao2023detrs}
Yian Zhao, Wenyu Lv, Shangliang Xu, Jinman Wei, Guanzhong Wang, Qingqing Dang, Yi Liu, and Jie Chen.
\newblock Detrs beat yolos on real-time object detection.
\newblock {\em arXiv preprint arXiv:2304.08069}, 2023.

\bibitem{zhu2020deformable}
Xizhou Zhu, Weijie Su, Lewei Lu, Bin Li, Xiaogang Wang, and Jifeng Dai.
\newblock Deformable detr: Deformable transformers for end-to-end object detection.
\newblock In {\em International Conference on Learning Representations}, 2021.

\bibitem{zohar2023prob}
Orr Zohar, Kuan-Chieh Wang, and Serena Yeung.
\newblock Prob: Probabilistic objectness for open world object detection.
\newblock In {\em Proceedings of the IEEE/CVF Conference on Computer Vision and Pattern Recognition}, pages 11444--11453, 2023.

\end{thebibliography}
}

\end{document}